\documentclass{article} 
\usepackage{iclr2026_conference,times}

\usepackage[utf8]{inputenc}
\usepackage[T1]{fontenc}
\usepackage{graphicx}
\usepackage{hyperref}
\usepackage{url}
\usepackage{amsfonts}
\usepackage{amsmath}
\usepackage{booktabs}
\usepackage[table]{xcolor}
\usepackage{subcaption}
\usepackage{wrapfig}
\usepackage{algorithm}
\usepackage{algorithmic}
\usepackage{microtype}

\title{Prompt Replay: Speeding up GRPO with On-Policy Reuse of High-Signal Prompts} 

\author{Andrei Baroian\thanks{Correspondence: \texttt{m.a.baroian@umail.leidenuniv.nl}} \qquad\qquad Rutger Berger \\
Leiden Institute of Advanced Computer Science\\
Leiden University
}

\iclrfinalcopy

\begin{document}
\maketitle
\lhead{}

\begin{abstract}
    Reinforcement learning with verifiable rewards (RLVR) plays a crucial role in expanding the capacities of LLM reasoning, but GRPO-style training is dominated by expensive rollouts and wastes compute on unusable prompts.  We propose \emph{Prompt Replay}, an overhead-free online data selection method for GRPO that reuses \emph{prompts only} (not trajectories), to preserve on-policy optimization.
    After each step, we insert prompts with medium difficulty into a buffer, and prioritize prompts closer to a pass rate of 0.5 (half answers correct, half wrong) to maximize the advantage, thus learning signal.
    Training batches are formed by mixing reused prompts with fresh samples, with cooldown steps and max reuse times controlling aggressiveness vs risk of overfitting.
    Across multiple model families (Llama-3.2-3B, Qwen3-8B) and training datasets (Dolci, Polaris), evaluated using average accuracy on six standard math benchmarks, Prompt Replay reduces zero-variance prompts, increases mean absolute advantage and shows faster initial accuracy gains. Yet, it plateaus and converges with the baseline, as too aggressive configuration was used. The method is most efficient when the rollouts are the primary bottleneck and the dataset is difficult for the model. We additionally observe that Qwen2.5-Math can exhibit spurious-reward effects that invalidates ablations, raising a warning signal for using it as a sole testbed for GRPO method research.

\end{abstract}

\section{Introduction}

Recently, Reinforcement Learning (RL) has emerged as a central technique within the fine-tuning paradigm of Large Language Models (LLMs). Various state of the art open-source methods, such as DeepSeek R1~\citep{Guo_2025}, have shown to push the LLM's reasoning capacities by incorporating RL with Verifiable Rewards (RLVR) inside post-training recipes \citep{olmo2025olmo3,team2025kimi, an2025polaris, lambert2025tulu3pushingfrontiers}. RLVR essentially uses binary rewards to grade correctness and has proven to be extremely powerful, enabling LLMs to achieve human-level capacities with regard to math olympiad questions \citep{liu2025prorl, luo2025deepscaler}.

Algorithmic implementations in RLVR moved from a \textit{PPO}-style actor-critic version towards \textit{GRPO} \citep{shao2024deepseekmathpushinglimitsmathematical}, \textit{Dr. GRPO} \citep{liu2025understandingr1zeroliketrainingcritical} and \textit{DAPO} \citep{yu2025dapo}, as to avoid the expensive critic model. However, with advantage calculated based on a group of multiple rollouts, rollouts are known to become an important bottleneck in the training stage, being responsible for the majority of computational costs~\citep{olmo2025olmo3}. Moreover, a lot of compute is wasted on prompts with zero variance (where answers are all correct or all wrong), which cannot be used for training. 
As a consequence, two directions emerged in the literature to optimize the expensive RL loop for LLMs: (i) predicting if a prompt has zero variance without performing all the rollouts, (ii) increasing sample efficiency by using prompts with higher learning signal.

\begin{figure}[t]
    \centering
    \includegraphics[width=1\linewidth]{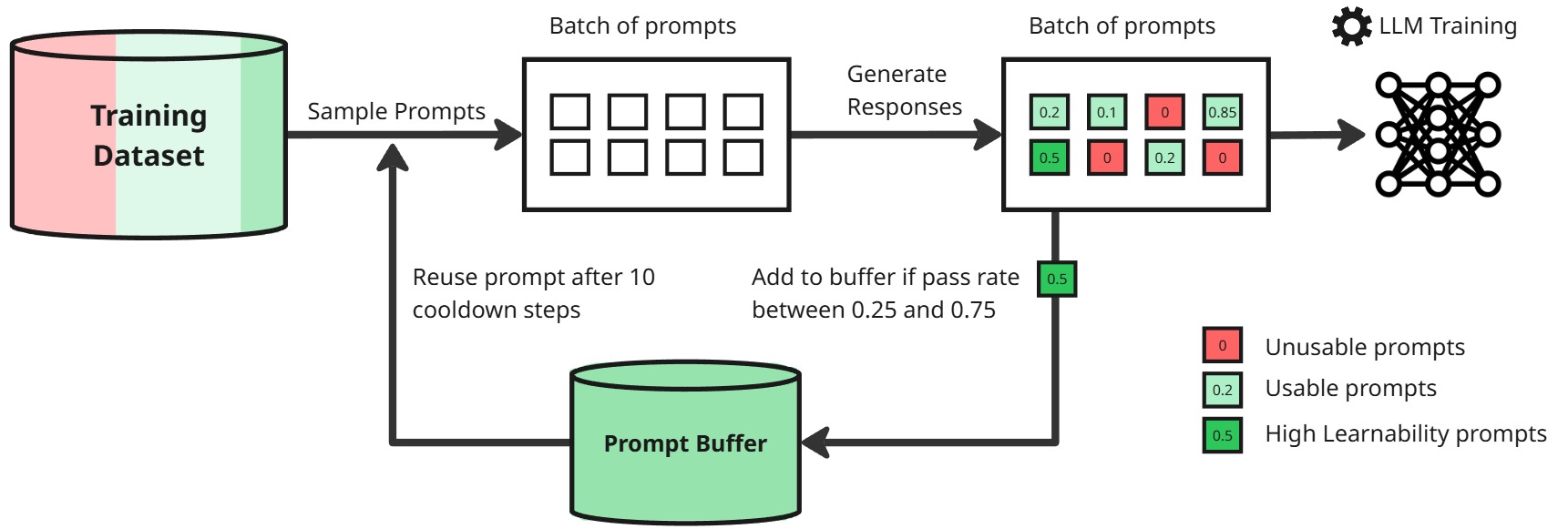}
    \caption{Prompt Replay visualization. In each step, we insert prompts with high learnability into a buffer. Training batches are formed by mixing reused prompts with fresh samples, with cooldown steps and max reuse times are controlling aggressiveness vs risk of overfitting.}
    \label{fig:placeholder}
    \vspace{-0.1in}
\end{figure}

Some works propose to store \textit{high learnable} off-policy trajectories inside a replay buffer~\citep{zhan2025exgrpolearningreasonexperience, zhang2025improvingsamplingefficiencyrlvr}. By replaying prompts with corresponding rollouts, the authors boost sample efficiency; however, they inherently introduce noise by including off-policy reasoning behaviours. 

Meanwhile, static data selection methods, such as \textit{LIMR} \citep{li2025limr} or \textit{s1} \citep{muennighoff2025s1simpletesttimescaling}, filter datasets \textit{offline}, based on predefined quality scores, to carefully select only the most useful questions. However, as the policy evolves, static metrics do not capture the dynamic nature of a model's capabilities. Consequently, much work has been put into \textit{Online Curriculum Learning} for RLVR \citep{shi2025efficient, zheng2025act, zeng2025cures}, where prompts are filtered based on current performance. 

Despite these advances, a significant challenge remains. Most online methods suffer from what can be called a ``measurement tax": to determine if a prompt is suitable, they often rely on redundant rollouts (rejection sampling) \citep{bae2025online, zhang2025speed}, external models~\citep{gao2025prompt}, or rather complicated (and therefore not very attractive) ways to estimate usefulness \citep{zeng2025cures}. Recently, GRESO \citep{zheng2025act} proposed a method to overcome this measurement tax, resulting in far fewer roll-outs required to achieve convergence. However, the authors leave sorting prompts on learnability for future research, and it relies on epoch-level statistics; one must wait for a full training epoch to update the data distribution. 
Moreover, none of these works combine sorting techniques (train from high to low utility) with predicting zero variance prompts.

In this work, we propose \textbf{Prompt Replay}, a dynamic data selection framework that addresses this gap in research. Following the intuition built in GRESO~\citep{zheng2025act}, we argue that an ideal selective rollout algorithm must obey two principles: 1) \textit{Online data selection} 
, and 2) \textit{Zero additional overhead} for difficulty estimation. Prompt Replay stores prompts only (without their responses) with high learnability in a buffer, and they are reused later in the training, balancing them with fresh prompts (global data). Inside the replay buffer, we sort prompts based on learnability, closer to a pass rate of 0.5 (half the answers were correct, half wrong). With this method, we significantly reduce the required number of rollouts and simultaneously increase the mean advantage per rollout, which leads to an increase in efficiency.

Our experiments cover different models (Qwen \citep{yang2025qwen3}, Llama \citep{llama32_3b_instruct_hf}), different sizes (1.5B, 3B, 8B), and different math datasets (Dolci RL-Zero Math \citep{olmo2025olmo3}, Polaris 53k \citep{Polaris2025}). We evaluate using the averaged accuracy on 6 common math benchmarks, and use the recently released OLMo-RL \citep{olmo2025olmo3} algorithm as a strong baseline.
Results show Prompt Replay dramatically reduces the number of zero-variance prompts, achieves a higher absolute mean advantage, and achieves high accuracy faster than the baseline, yet, it eventually plateaus and converges with the baseline. We test different cooldown steps on Qwen 2.5 Math 1.5B \citep{yang2024qwen25math} and find out that applying our method on this model is unreliable, as it does not generalize to other model families. 

Our main contributions can be summarized as follows:
\begin{itemize}

\item \textbf{Prompt Replay: an Overhead-Free Online Filtering method.} We propose a 
data selection framework that uses a Replay Buffer to prioritize prompts with the highest theoretical learning signal. It significantly reduces the required number of rollouts and simultaneously increases the mean advantage, with zero additional overhead.

\item \textbf{Robust Generalization and Efficiency.} We validate our method across \emph{multiple training datasets} and \emph{three distinct models} and a diverse set of benchmarks. We highlight that Prompt Replay's advantages shine in scenarios where rollouts are the bottleneck and the dataset is difficult for the model at hand. 
\item  \textbf{We raise a warning signal against using Qwen 2.5 Math} model family, which has become close to a standard in the GRPO literature, as it provides a false positive signal for researchers, by gaining easy improvements in accuracy, but has poor generalization to other models.

\end{itemize}

\section{Preliminaries}

\subsection*{Group Relative Policy Optimization (GRPO)}

\label{sec:grpo}

GRPO generates a group of $G$ candidate responses $\{y_j\}_{j=1}^{G}$ (i.e., performs rollouts) for each prompt $x$ (question) using the LLM as policy. The response is checked against the ground-truth answer provided by the dataset, and a binary reward is assigned: $r(x,y)=1$ if the final answer matches the ground truth, and $r(x,y)=0$ otherwise. It then assigns each response a \emph{group-relative} advantage by subtracting from its reward the mean of the other responses for the same prompt.

Building upon the original Group Relative Policy Optimization (GRPO) algorithm~\citep{shao2024deepseekmathpushinglimitsmathematical}, much effort has been put into optimizing the GRPO framework~\citep{shao2024deepseekmathpushinglimitsmathematical, yu2025dapo, liu2025understandingr1zeroliketrainingcritical}. Our method builds upon the formulation of \textit{OLMo-RL}~\citep{olmo2025olmo3}, which includes most of the latest established advancements. 

Formally, the objective function is defined as:
\begin{equation}
    \label{eq:olmo_grpo}
    \begin{split}
        J(\theta) = \frac{1}{\sum_{i=1}^{G} T_i} 
        \sum_{i=1}^{G} \sum_{t=1}^{T_i} 
        & \min \left( \frac{\pi_{\theta}(y_{i,t} \mid x, y_{i,<t})}{\pi_{\theta_{\text{old}}}(y_{i,t} \mid x, y_{i,<t})}, \eta \right) \\
        & \cdot \min \left( \rho_{i,t}(\theta) \hat{A}_{i}, \operatorname{clip}(\rho_{i,t}(\theta), 1 - \epsilon_{\text{low}}, 1 + \epsilon_{\text{high}}) \hat{A}_{i} \right),
    \end{split}
\end{equation}
where $T_i$ is the length of response $y_i$. The term $\rho_{i,t}(\theta) = \frac{\pi_{\theta}(y_{i,t}\mid x,y_{i,<t})}{\pi_{\theta_{\text{old}}}(y_{i,t}\mid x,y_{i,<t})}$ represents the token-level probability ratio between the current and old policies. The hyperparameters $\epsilon_{\text{low}}$ and $\epsilon_{\text{high}}$ control the clipping range, while $\eta$ serves as a truncated importance sampling cap~\citep{yao2025your}.

The responses $y_i$ are sampled from the old policy $y_i \sim \pi_{\theta_{\text{old}}}(\cdot \mid x)$. Consistent with the GRPO formulation, the advantage $\hat{A}_{i}$ applies to all tokens $t$ in response $y_i$ and is calculated based on the relative reward within the group:
\begin{equation}
    \hat{A}_{i} = r(x, y_i) - \operatorname{mean}\left( \{ r(x, y_j) \}_{j=1}^G \right)
    \label{eq:advantage_centered}
\end{equation}
where $x$ denotes the input prompt, $y_i = (y_{i,1}, \dots, y_{i,T_i})$ is the generated response sequence, and $r(x, y_i)$ is the scalar reward assigned to response $y_i$ for prompt $x$.
Note that this variant uses mean-centering without standard deviation normalization.

\subsection*{Pass Rate and Theoretical Efficiency of Gradient Updates.}
\label{sec:theory-efficiency}
In the RL framework, we can define training efficiency not strictly by the number of update steps, but rather by the magnitude of loss reduction per optimization step. It is shown that this efficiency is bounded by the variance of the reward signal and maximizes at medium difficulty questions, where half of the responses are correct (pass rate, or $p_\theta(x)=0.5$)~\citep{zeng2025cures, foster2025learningreasonfrontierlearnability, bae2025online}. A pass rate of 0.5 creates the highest mean absolute advantage and subsequently the highest gradient signal. Prior work demonstrated this on slightly different objectives and advantage; we write the derivation with the used objective in Appendix \ref{appendix:px_theory}.

\section{Related Work}

\paragraph{RL for LLM Reasoning} 
Recently, Reinforcement Learning (RL) has evolved from human preference alignment ~\citep{christiano2023deepreinforcementlearninghuman, ouyang2022traininglanguagemodelsfollow, bai2022traininghelpfulharmlessassistant} to Reinforcement Learning with Verifiable Rewards (RLVR) for reasoning tasks \citep{Guo_2025, yu2025dapo}. While early RLVR methods relied on Proximal Policy Optimization (PPO) and (expensive) value models, advances such as Group Relative Policy Optimization (GRPO) \citep{shao2024deepseekmathpushinglimitsmathematical} have significantly reduced computational overhead by proposing group-based advantage estimation. Subsequent optimizations, including \textit{Dr. GRPO} \citep{liu2025understandingr1zeroliketrainingcritical}, \textit{DAPO} \citep{yu2025dapo}, or \textit{OLMO 3} \citep{olmo2025olmo3}, further refine efficiency through importance sampling truncation, length-bias mitigation, and the removal of the KL penalty. Our work builds upon these algorithmic refinements, following most recent line of research~\citep{olmo2025olmo3}.

\paragraph{Data Selection and Prompt Utility}
Besides algorithmic changes, data selection strategies aim to optimize efficiency by training on the most informative samples. Essentially, many prompts remain unusable for many epochs, being either too difficult or too easy, delivering limited training signals. Offline selection methods like \textit{LIMR} \citep{li2025limr} and \textit{s1} \citep{muennighoff2025s1simpletesttimescaling} prune datasets based on static quality scores. However, as the policy evolves, static selection fails to capture the dynamic nature of the model's capacity. Recent work defines \textit{prompt utility} through various lenses: signal-to-noise maximization \citep{zhang2025speed}, maximal gradient norms \citep{zeng2025cures}, or estimated learning impact \citep{li2025limr, foster2025learningreasonfrontierlearnability}. Independent of the lens looked through, it is shown that the most useful prompts are of medium difficulty, where half of the rollouts obtain a reward (i.e. half the answers are correct, half wrong).

\paragraph{Curriculum Learning and Online Filtering}
Curriculum learning (CL) organizes tasks from easy to hard to improve convergence \citep{bengio2009curriculum}. In the context of LLMs, static curricula \citep{luo2025deepscaler, song2025fastcurlcurriculumreinforcementlearning} use hand-crafted difficulty schedules, while online strategies \citep{shi2025efficient, gao2025prompt, zheng2025act} adaptively filter prompts based on the model's current performance. However, a significant bottleneck remains: many online methods \citep{bae2025online,zhang2025speed}, rely on additional rollouts to estimate difficulty, introducing computational overhead. In other work~\citep{zhang2025speed}, redundant models are introduced to perform the same task. Methods like \textit{CurES} \citep{zeng2025cures} use Bayesian estimation to mitigate this, but still require complex scheduling.

\paragraph{Experience Replay and Stability}
Experience Replay (ER) is a reinforcement learning technique originally proposed to stabilize training and improve sample efficiency by storing and re-sampling previously generated experiences \citep{lin1992self}. While on-policy algorithms like GRPO typically discard rollouts after a single update, recent works have adapted ER for LLMs to optimize the training loop, showing notable improvements in convergence speed and learning capabilities \citep{zhan2025exgrpolearningreasonexperience, zhang2025rlepreinforcementlearningexperience, dou2025improvingrlexplorationllm}. However, integrating ER into on-policy LLM training introduces two new challenges. First, defining which data to keep inside the buffer is non-trivial \citep{zhan2025exgrpolearningreasonexperience, schaul2016prioritizedexperiencereplay}. Second, standard ER implementations often store full trajectories (prompts and responses). While this increases sample efficiency \citep{qu2025can, sun2025improving, zhang2025improvingsamplingefficiencyrlvr}, it introduces off-policy noise from previous, less capable model behaviours. Furthermore, aggressive reuse without regularization can lead to catastrophic forgetting, particularly when the KL divergence penalty is removed. To mitigate these concerns, we replay \textit{only prompts} to maintain on-policy optimization and mix these with global (fresh) data samples.

\section{Methodology}

\subsection{Prompt Replay}

To maximize sample efficiency, we maintain a dynamic replay buffer $\mathcal{B}$.
After each training step, we estimate the pass rate $p_\theta(x)$ for each sampled prompt $x$.
Prompts with intermediate difficulty, $p_\theta(x)\in[p_{\min},p_{\max}]$, are inserted into the buffer.

At step $t$, we build a batch of size $N$ by mixing \emph{fresh} prompts from the original dataset with prompts drawn from the buffer.
We reuse up to a fraction $\epsilon$ of the batch, but the realized fraction $\epsilon_t\le \epsilon$ depends on how many buffer prompts are
\emph{eligible} (a prompt is eligible if its cooldown has expired).
This yields the mixed prompt-sampling distribution
\begin{equation}
D_{\text{mix},t}
\;=\;
(1-\epsilon_t)\,D_{\text{fresh}}
\;+\;
\epsilon_t\,D_{\text{buf},t},
\label{eq:promptreplay_mix}
\end{equation}
where $D_{\text{fresh}}$ is the default sampling distribution over the full dataset (i.e., the standard sampler used without replay),
and $D_{\text{buf},t}$ is the distribution induced by sampling eligible prompts from $\mathcal{B}_t$.

Prompt Replay is \emph{on-policy}: buffer entries store only prompts (no trajectories), and for each reused prompt we generate new
completions using the current policy.
Within $\mathcal{B}_t$, we prioritize medium-difficulty prompts by ranking eligible prompts by
$|p_\theta(x)-0.5|$ and sampling from the top of this ordering (ties sampled uniformly).

When a prompt is inserted into $\mathcal{B}$, a cooldown period $C$ prevents it from being reused for the next $C$ training steps.
Each prompt can be reused at most $R$ times; after $R$ reuses it is removed from $\mathcal{B}_t$ but remains available via the
standard sampler $D_{\text{fresh}}$, preventing depletion in long multi-epoch runs.
After each reuse, we recompute $p_\theta(x)$ under the current policy and reinsert the prompt into $\mathcal{B}$ only if
$p_\theta(x)\in[p_{\min},p_{\max}]$. Full pseudo code can be found in Appendix \ref{alg:grpo_prompt_replay}.

The hyperparameters $(\epsilon,C,R)$ trade off sample-efficiency gains against
overfitting risk: larger $\epsilon$ and $R$ and smaller $C$ increase reuse concentration on a small set of prompts.

Prompt Replay is a similar concept to Prioritized Experience Replay~\citep{schaul2016prioritizedexperiencereplay}, except that it does not store the trajectories (answers), only the prompts (questions), and it prioritizes based on pass rate rather than TD error. 

\subsection{Theoretical Justification} \label{sec:active-sampling-theory}

We justify our sampling strategy by viewing the training process as a \textit{constrained resource allocation problem}. We aim to maximize the expected policy improvement per optimization step while ensuring the coverage of the support with the global data distribution.

Let $\mathcal{D}$ be the global dataset and $B$ be the fixed batch size for a training step. We seek a subset $\mathcal{B} \subset \mathcal{D}$ with $|\mathcal{B}| = B$ that maximizes the expected gradient norm, which serves as a proxy for learning speed.

As shown in Appendix~\ref{appendix:px_theory}, the contribution of a single sample $x$ to the gradient variance is proportional to the variance of its reward distribution: 
\begin{equation} 
\mathbb{E} [ \| \nabla_\theta J(x) \|^2 ] \propto \operatorname{Var}(r|x) = p_\theta(x)(1 - p_\theta(x)). 
\end{equation} 

Let $v(x) = p_\theta(x)(1 - p_\theta(x))$ denote this value function. The function $v(x)$ is strictly concave over $p_\theta(x) \in [0,1]$ and symmetric around its global maximum at $p_\theta(x) = 0.5$.

To maximize the total learning signal of the batch, we solve: 
\begin{equation} 
\max_{\mathcal{B} \subset \mathcal{D}} \sum_{x \in \mathcal{B}} v(x), \quad \text{s.t. } |\mathcal{B}| = B. 
\end{equation} 

Since $v(x)$ is monotonically decreasing with respect to the distance $\delta(x) = |p_\theta(x) - 0.5|$, the optimal solution to this maximization problem is the set of $B$ samples with the smallest $\delta(x)$. Our sorting mechanism, prioritizing prompts by $|p_\theta(x) - 0.5|$ is equivalent to a greedy maximization of the expected gradient magnitude per batch.


While sorting maximizes the update magnitude, sampling exclusively from the frontier subset $\mathcal{S}$ (defined as the set of prompts with minimal $\delta(x)$) introduces a distributional shift. The true optimization objective is the expectation over the global distribution $\mathcal{D}$: 
\begin{equation} 
J_{\text{true}}(\theta) = \mathbb{E}_{x \sim \mathcal{D}} [J(x; \theta)]. 
\end{equation} 

If we optimize solely on the active subset $\mathcal{S}$, we are optimizing a different objective $J_{\mathcal{S}}(\theta) = \mathbb{E}_{x \sim {\mathcal{S}}} [J(x; \theta)]$. Crucially, the gradient $\nabla J_{\mathcal{S}}$ has zero support on regions where $x \notin \mathcal{S}$ (i.e., easy or impossible prompts). This leads to two failure modes: 

\begin{enumerate} 
    \item \textbf{Catastrophic Forgetting:} The model may degrade on previously solved tasks ($p_\theta(x) \approx 1$) because the active objective $J_{\mathcal{S}}$ imposes no penalty for such degradation. 
    \item \textbf{Lack of Exploration:} By excluding difficult tasks ($p_\theta(x) \approx 0$), we could prevent them from ever transitioning into the ``learnable'' region ($p_\theta(x) \approx 0.5$) as the model evolves. 
\end{enumerate}

To address this, we define a mixture distribution $Q(x) = \epsilon {\mathcal{S}}(x) + (1 - \epsilon) {\mathcal{D}}(x)$. This ensures that the support of the sampling distribution covers the support of the true distribution ($\text{supp}({\mathcal{D}}) \subseteq \text{supp}(Q)$).

The expected gradient under our mixture strategy is: 
\begin{equation} 
\mathbb{E}_{x \sim Q} [\nabla J(x)] = \epsilon \underbrace{\mathbb{E}_{x \sim {\mathcal{S}}} [\nabla J(x)]}_{\text{High Variance (Speed)}} + (1-\epsilon) \underbrace{\mathbb{E}_{x \sim {\mathcal{D}}} [\nabla J(x)]}_{\text{Global Anchor (Stability)}}. 
\end{equation}

The term $(1 - \epsilon) \mathbb{E}_{x \sim {\mathcal{D}}} [\nabla J(x)]$ acts as an approximation of the global constraints. Even if the gradient magnitude on easy tasks is small, it could provide necessary directional information to prevent the policy from drifting away from established knowledge. The coefficient $\epsilon$ controls the trade-off between the \textit{rate} of convergence and the \textit{stability} of the estimator.

\subsection{Experimental Setup}

OLMo-RL training codebase was used, providing a strong a baseline \citep{olmo2025olmo3}.

For \textbf{evaluation}, we report average accuracy over 6 benchmarks:
AIME25 \citep{aime25}, AIME24 \citep{aime24}, AMC \citep{amc23}, MATH500 \citep{math500},
OlympiadBench \citep{olympiadbench}, and MinervaMath \citep{minervamath}, all common practice used in the literature. We acknowledge the variance of some of them (AIME24, AIME25, AMC), as they contain a small number of data points (30, 30, and 40, respectively), yet using the average over multiple benchmarks ensures robustness of the evaluation.

For the \textbf{models}, we experiment with Qwen3-8B \citep{yang2025qwen3}, and Llama 3.2 3B \citep{llama32_3b_instruct_hf}. The method must work regardless of what dataset is chosen, so for \textbf{training datasets}, we use Dolci RL Zero Math from OLMo 3, \citep{olmo2025olmo3}, which contains 13k examples, and Polaris, a more difficult dataset with 53k examples \citep{Polaris2025} used only for the bigger Qwen3 8B model. These were chosen so they would vary in size and difficulty, and be appropriate for the model sizes chosen.

For experiments on the Dolci dataset, we set the total \textbf{context length} at 8192 tokens: 1024 for the prompt (question) and 7168 tokens for the response. For the Polaris dataset, we extend the context to 12,288 tokens total context with the same tokens allocated for the prompt. 

For \textbf{hardware}, we use 2 nodes of 4 H100 95GB each, one node for learners (weight update) and one for the rollouts (vLLM engines; actors), except for the Dolci dataset, where the context is too big and gets CUDA OOM, where we added another node for the rollouts, having two GPUs per vLLM engine.

For \textbf{hyperparameter tuning}, we test the cooldown steps $C=[2,5,10,20]$ on Qwen 2.5 Math 1.5B \citep{yang2024qwen25math} on the Dolci dataset with a total context of 4092, on a single node with 4 H100s, divided equally between learners and actors.

For Prompt Replay main results, we used $C=10$ cooldown steps, $R=15$ max reuse times, $p_{min} = 0.25, p_{max}=0.75$ and with prompt replay fraction  $\epsilon=0.75$. 

Learning rate is constant at 1e-6, sampling temperature set to 1 for rollouts and set to 0 for evaluation. We use a batch size of 32 prompts with 16 rollouts per prompt, resulting in 512 rollouts per training step. 
Full hyperparameter list in Appendix \ref{HPs}.

\section{Results}

\subsection{Main Results}

\begin{figure}[bt]
\centering

\hspace{-0.1\linewidth}
\makebox[\linewidth][c]{%
\begin{minipage}{1.15\linewidth}
\centering

\begin{minipage}[b]{0.06\linewidth}\end{minipage}%
\begin{minipage}[b]{0.31\linewidth}\centering Llama 3.1 4B (Dolci)\end{minipage}%
\begin{minipage}[b]{0.31\linewidth}\centering Qwen 3 8B (Dolci)\end{minipage}%
\begin{minipage}[b]{0.31\linewidth}\centering Qwen 3 8B (Polaris)\end{minipage}

\vspace{0.3em}

\begin{minipage}[c]{0.06\linewidth}\centering
  \raisebox{0.5\height}{\rotatebox{90}{\normalsize Accuracy vs time}}
\end{minipage}%
\begin{minipage}[c]{0.31\linewidth}\centering
  \includegraphics[width=\linewidth]{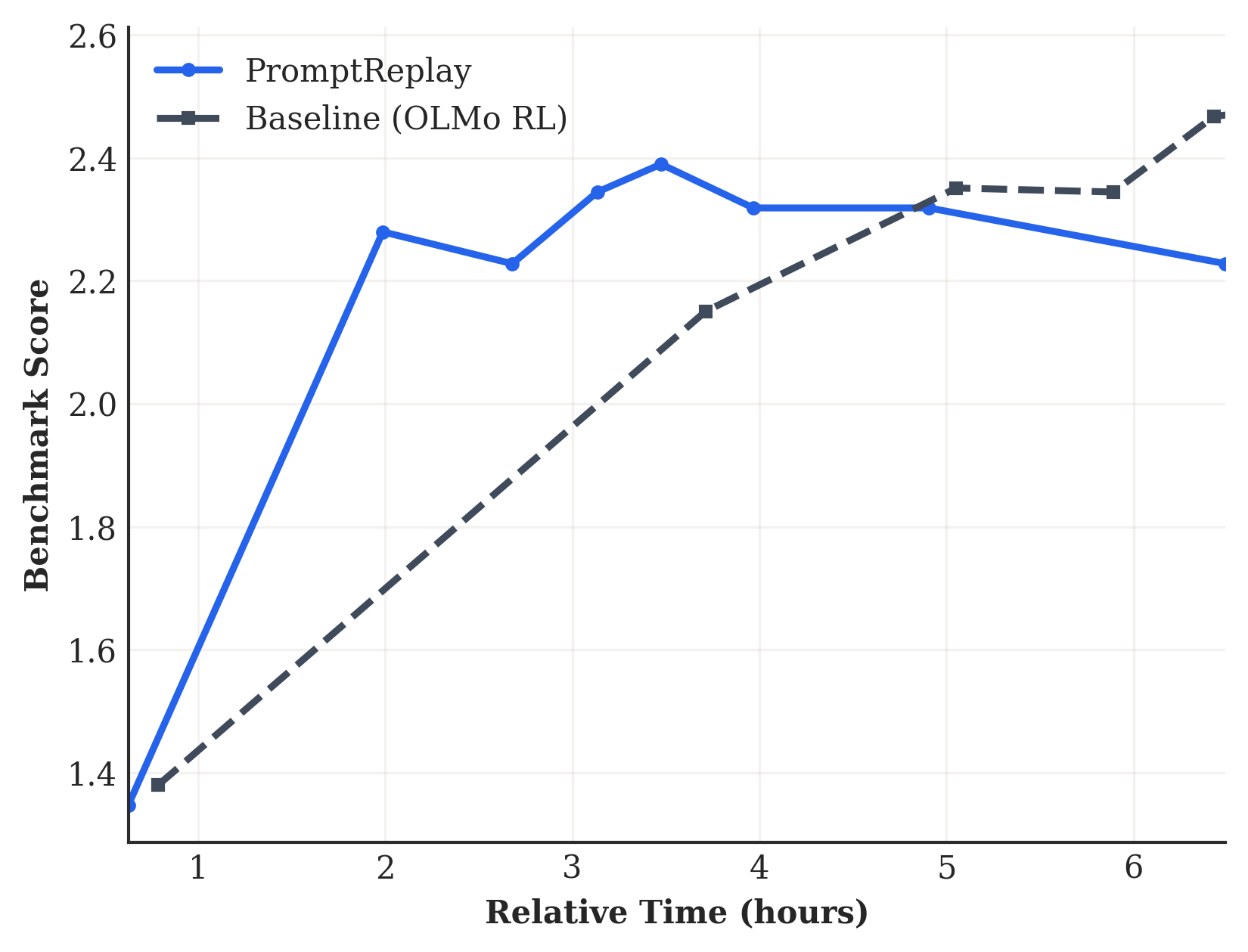}
\end{minipage}%
\begin{minipage}[c]{0.31\linewidth}\centering
  \includegraphics[width=\linewidth]{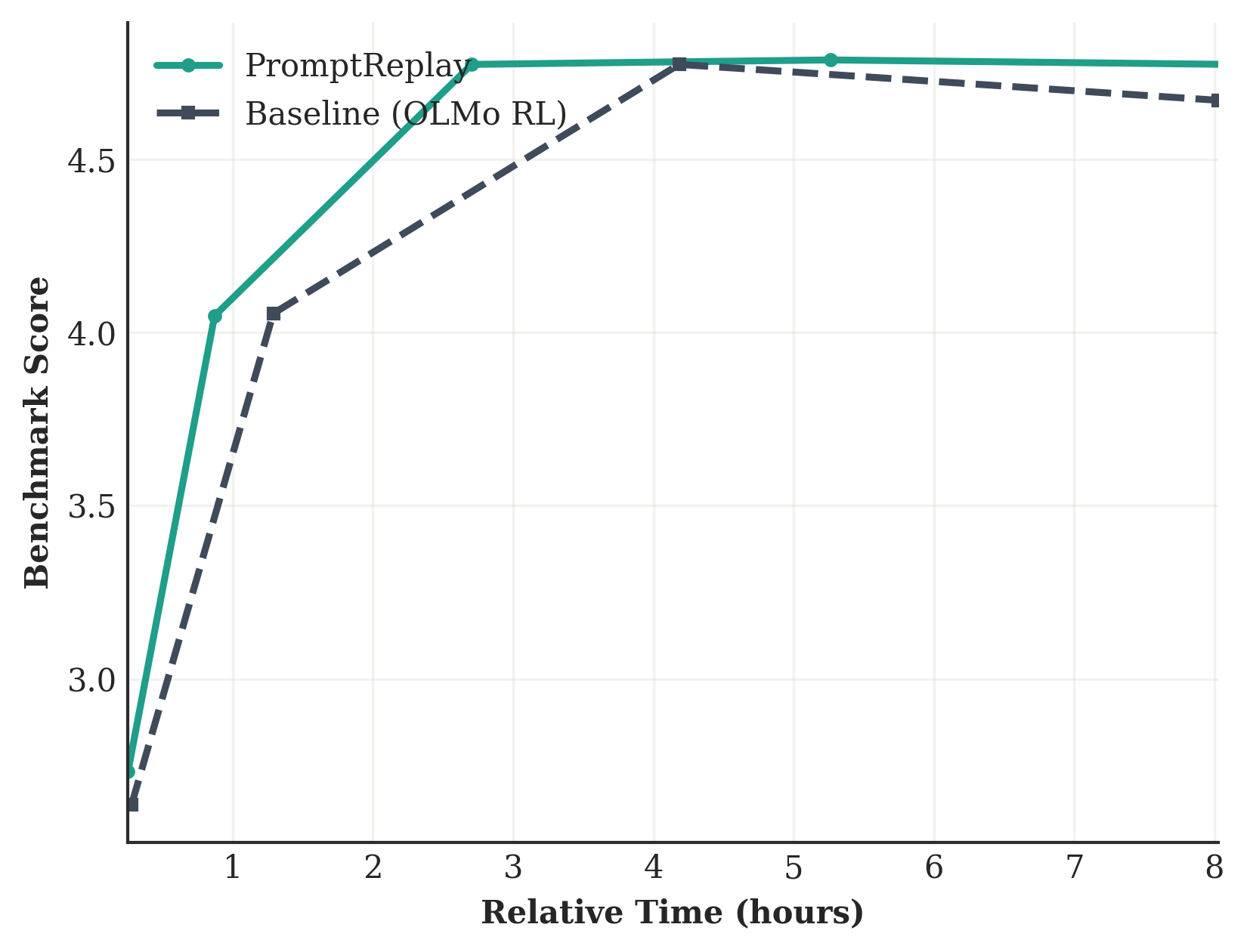}
\end{minipage}%
\begin{minipage}[c]{0.31\linewidth}\centering
  \includegraphics[width=\linewidth]{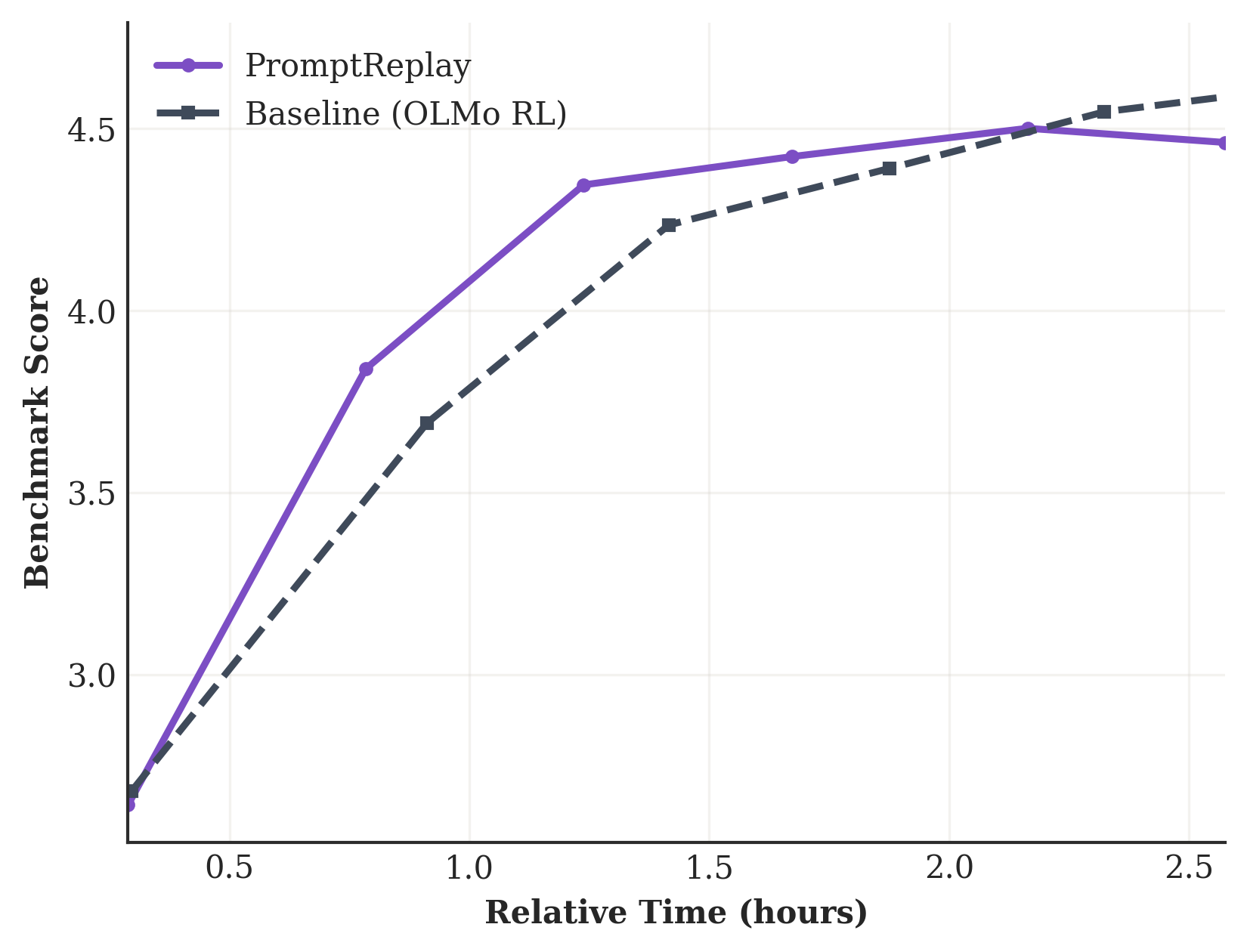}
\end{minipage}

\vspace{0.6em}

\begin{minipage}[c]{0.06\linewidth}\centering
  \raisebox{0.5\height}{\rotatebox{90}{\scriptsize No. of prompts px=0}}
\end{minipage}%
\begin{minipage}[c]{0.31\linewidth}\centering
  \includegraphics[width=\linewidth]{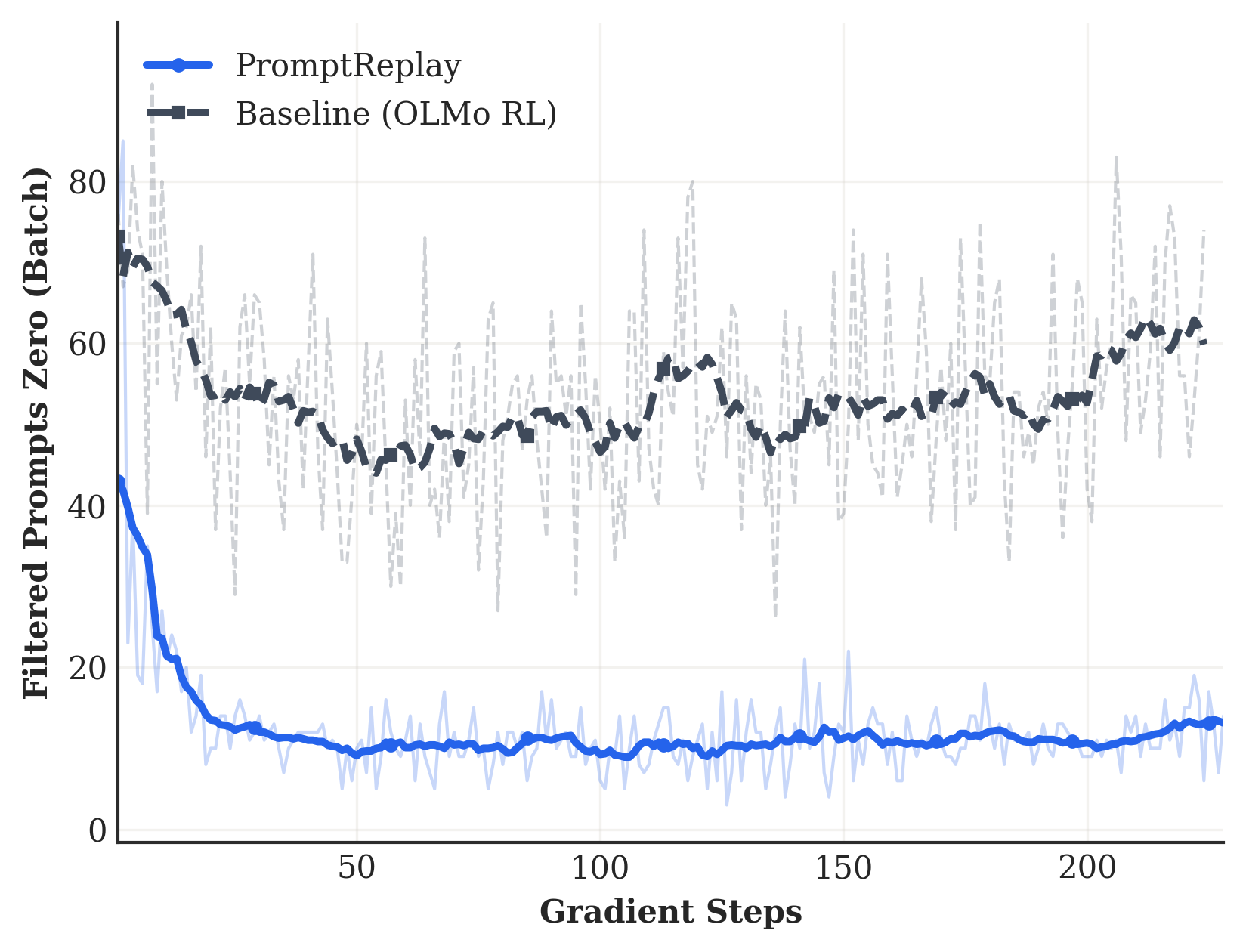}
\end{minipage}%
\begin{minipage}[c]{0.31\linewidth}\centering
  \includegraphics[width=\linewidth]{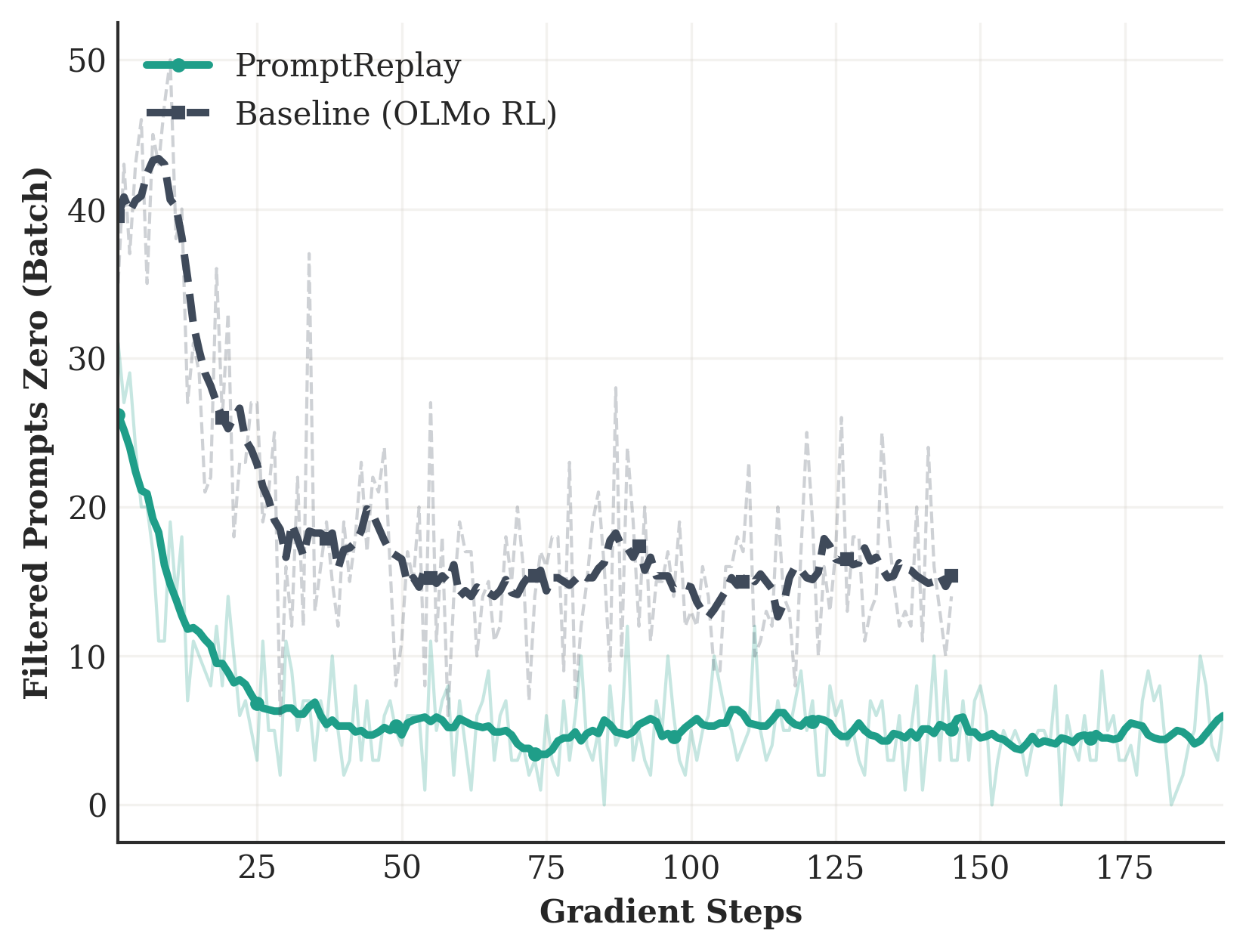}
\end{minipage}%
\begin{minipage}[c]{0.31\linewidth}\centering
  \includegraphics[width=\linewidth]{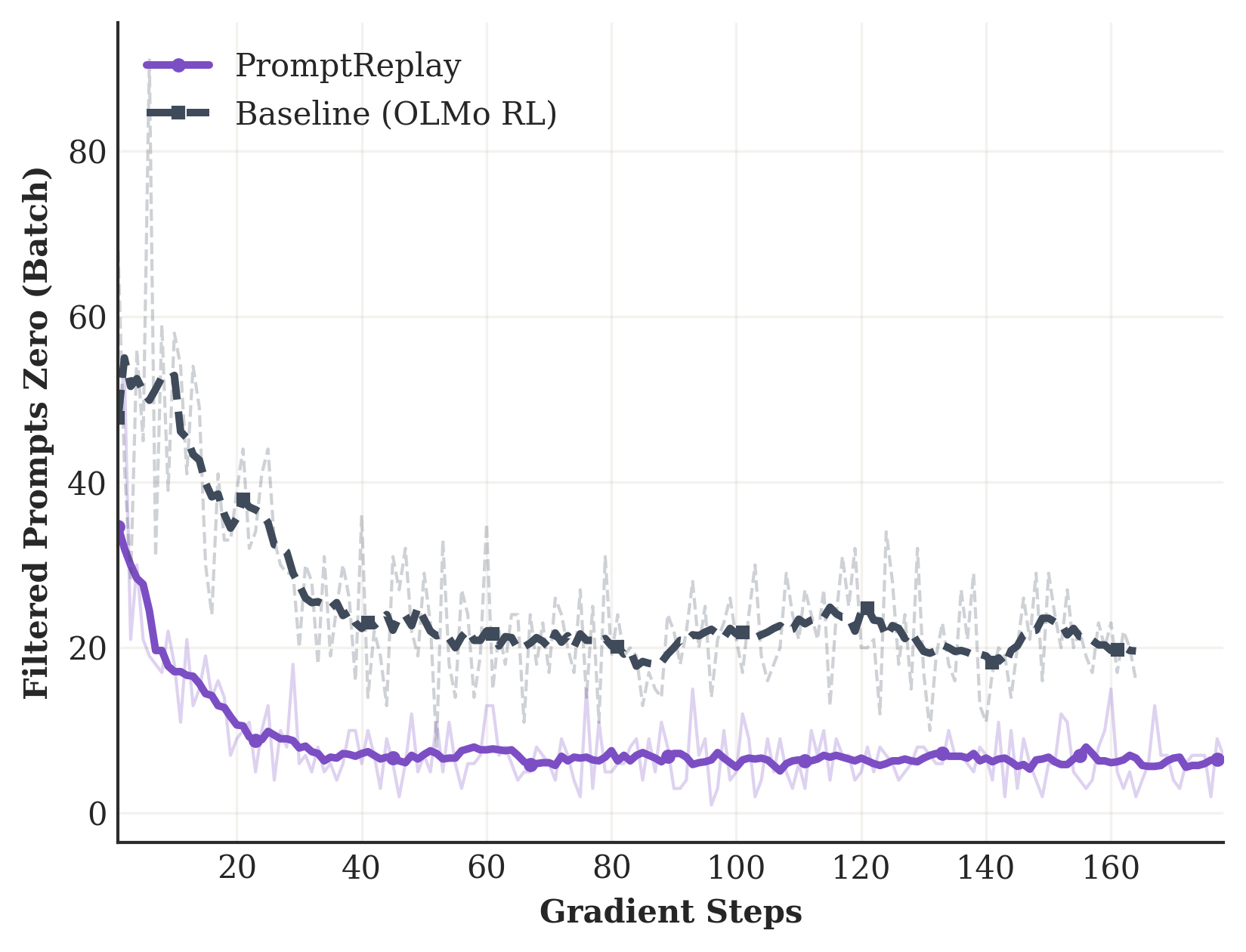}
\end{minipage}

\vspace{0.6em}

\begin{minipage}[c]{0.06\linewidth}\centering
  \raisebox{0.5\height}{\rotatebox{90}{\scriptsize Mean $|\mathrm{Adv}|$}}
\end{minipage}%
\begin{minipage}[c]{0.31\linewidth}\centering
  \includegraphics[width=\linewidth]{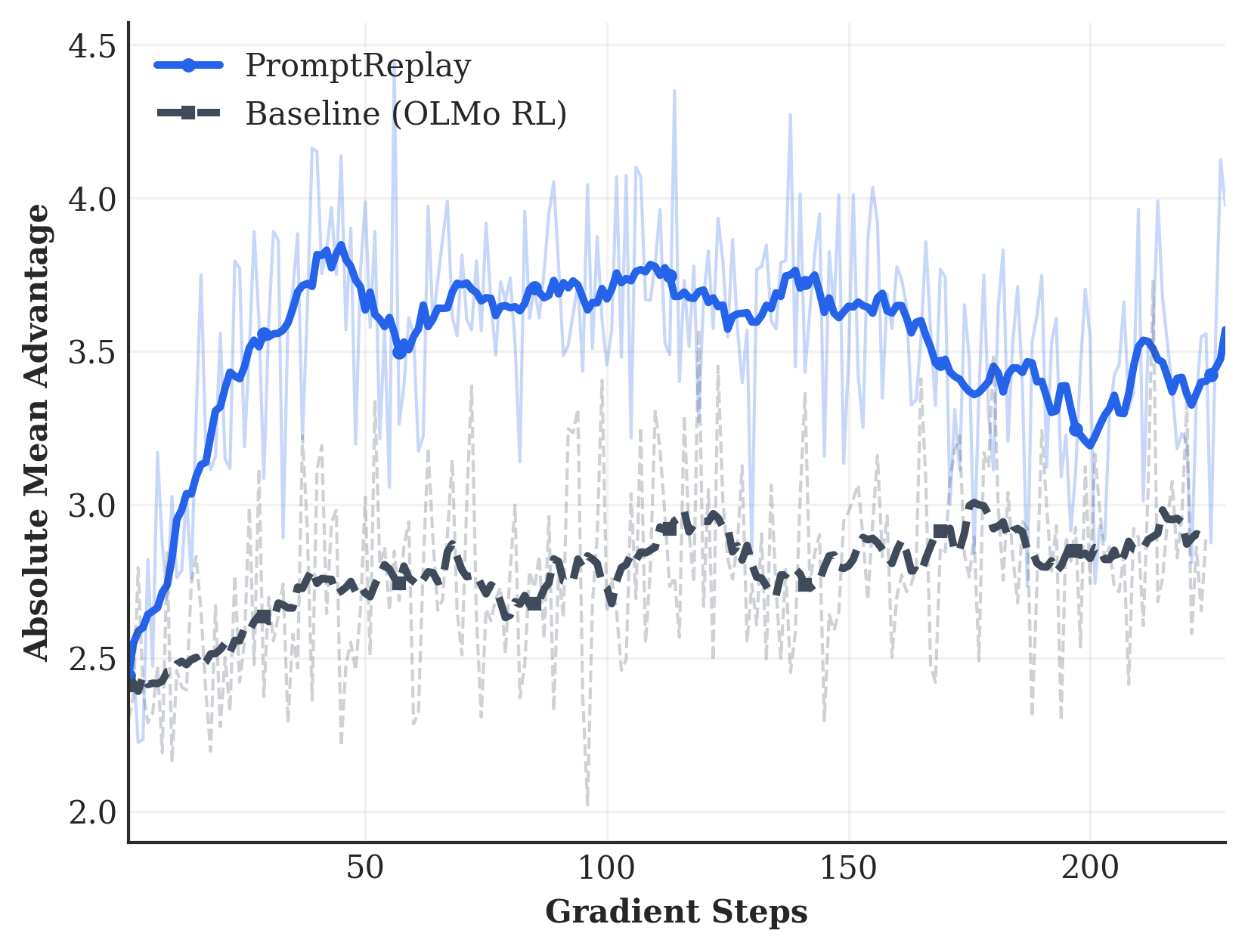}
\end{minipage}%
\begin{minipage}[c]{0.31\linewidth}\centering
  \includegraphics[width=\linewidth]{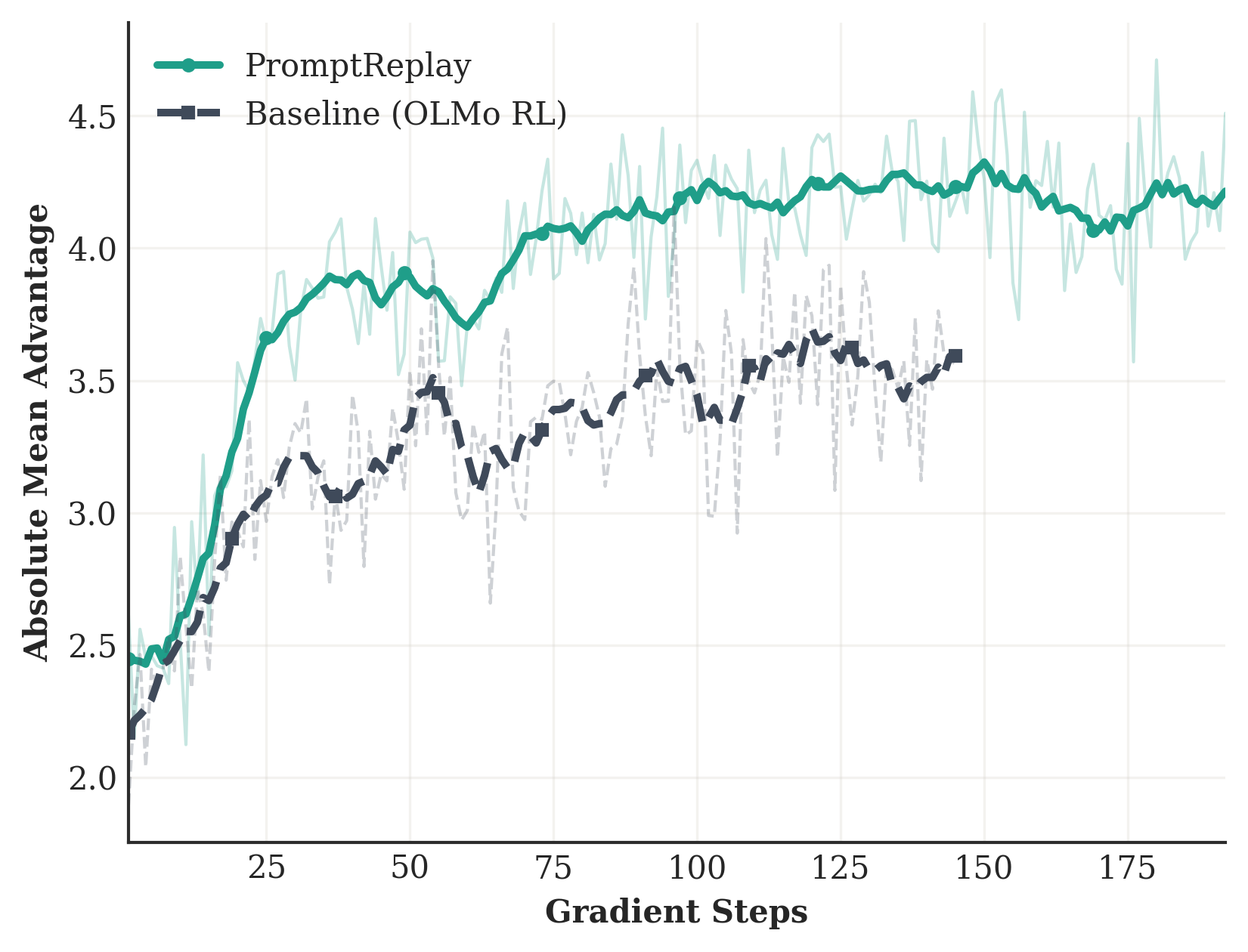}
\end{minipage}%
\begin{minipage}[c]{0.31\linewidth}\centering
  \includegraphics[width=\linewidth]{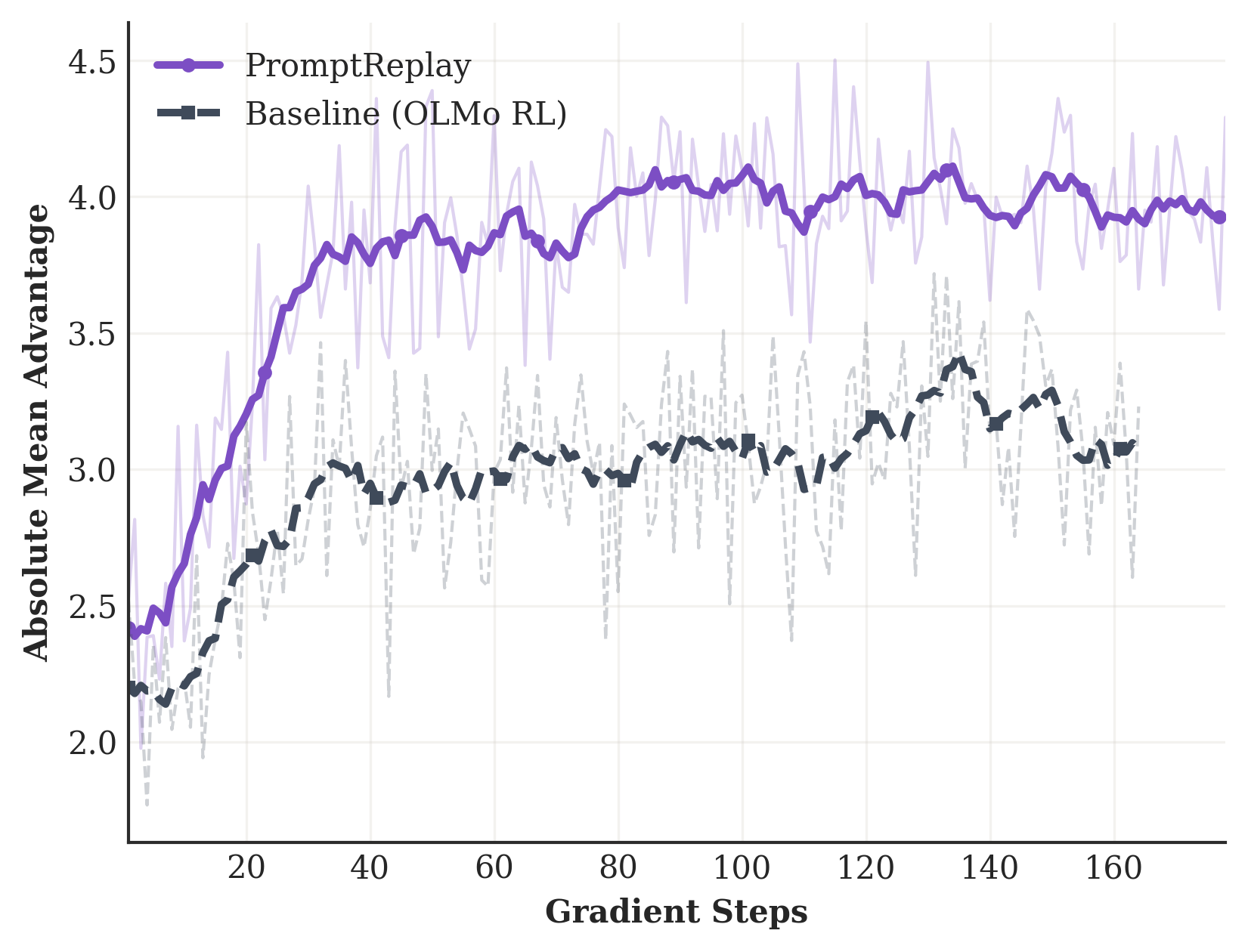}
\end{minipage}
\end{minipage}%
}

\caption{\textbf{Prompt Replay vs Baseline (OLMo-RL)}. Prompt Replay shows: higher mean $|\mathrm{Adv}|$ (3rd row), resulting in more signal from the data; lower number of prompts (2nd row) with pass rate = 0, wasting less compute on unusable prompts; earlier gains in the average accuracy over 6 benchmarks (1st row), but plateaus and converges with the baseline; benchmark score is max at 10 instead of 100}
\label{fig:models_comparison_9}

\vspace{-1pc}
\end{figure}

Figure \ref{fig:models_comparison_9} compares Prompt Replay with the baseline OLMo-RL: the first row represents the average accuracy of the 6 benchmarks, the second row represents the number of prompts with the pass rate equal to 0, and third row represents the mean absolute advantage. Prompt Replay works as expected, where it has a higher mean absolute advantage from reusing the prompts that are closer to a pass rate equal to 0.5 and saves compute by not performing rollouts on prompts that end up having pass rate of 0, thus performing more steps in the same amount of time (see Appendix \ref{fig:sup_results} )  for figure with training steps over training time). These translate into faster gains in accuracy on the benchmarks early on, but eventually plateau and converge with the baseline. 

The plateau of the prompt replay can be explained by two factors: (i) the hyperparameter configuration chosen might be too aggressive - it might reuse prompts too many times or too often - which possibly led to overfitting over a small number of prompts and stagnation of the accuracy; small experiments were run to test different configurations, yet the search space is large so the optimal might not have been found.  An extensive hyperparameter optimization remains for future work. (ii) the baselines plateau as well; for Qwen 3 8B Dolci, it is clear from the figure \ref{fig:models_comparison_9}, while for the other two experiments, the baseline was run for longer and showed the same plateau roughly 2\% points higher. So the plateau behaviour is expected, but occurs earlier for prompt replay due to aggressive hyperparameter choice.

There are two reasons why Qwen 3 8B models show smaller initial improvements compared to the Llama: (i) one of the benefits of prompt replay is that it saves up compute (and speeds up training) by reducing the rollouts done on those unusable prompts; Qwen is a bigger model, so on the same training dataset, it has a lower number of prompts with a pass rate of 0 meaning lower gains from prompt replay; this can be seen in Fig \ref{fig:models_comparison_9} second row, as the gap between baseline and prompt replay. (ii) On Qwen3-8B Polaris, the longer context forced us to double rollout GPUs (4→8; two GPUs per vLLM engine), so rollouts stopped being the bottleneck and Prompt Replay’s compute savings from fewer zero-variance prompts didn’t translate into faster training; the rollout GPUs were partially idle during weight updates.

These results show that Prompt Replay benefits are higher when (a) rollouts (generating the answer) are the bottleneck and (b) the dataset is difficult for the model (i.e. the fraction of prompts where model did not answer correct is high). 

\begin{figure}
    \centering

    \begin{subfigure}[t]{0.45\linewidth}
        \centering
        \includegraphics[width=\linewidth]{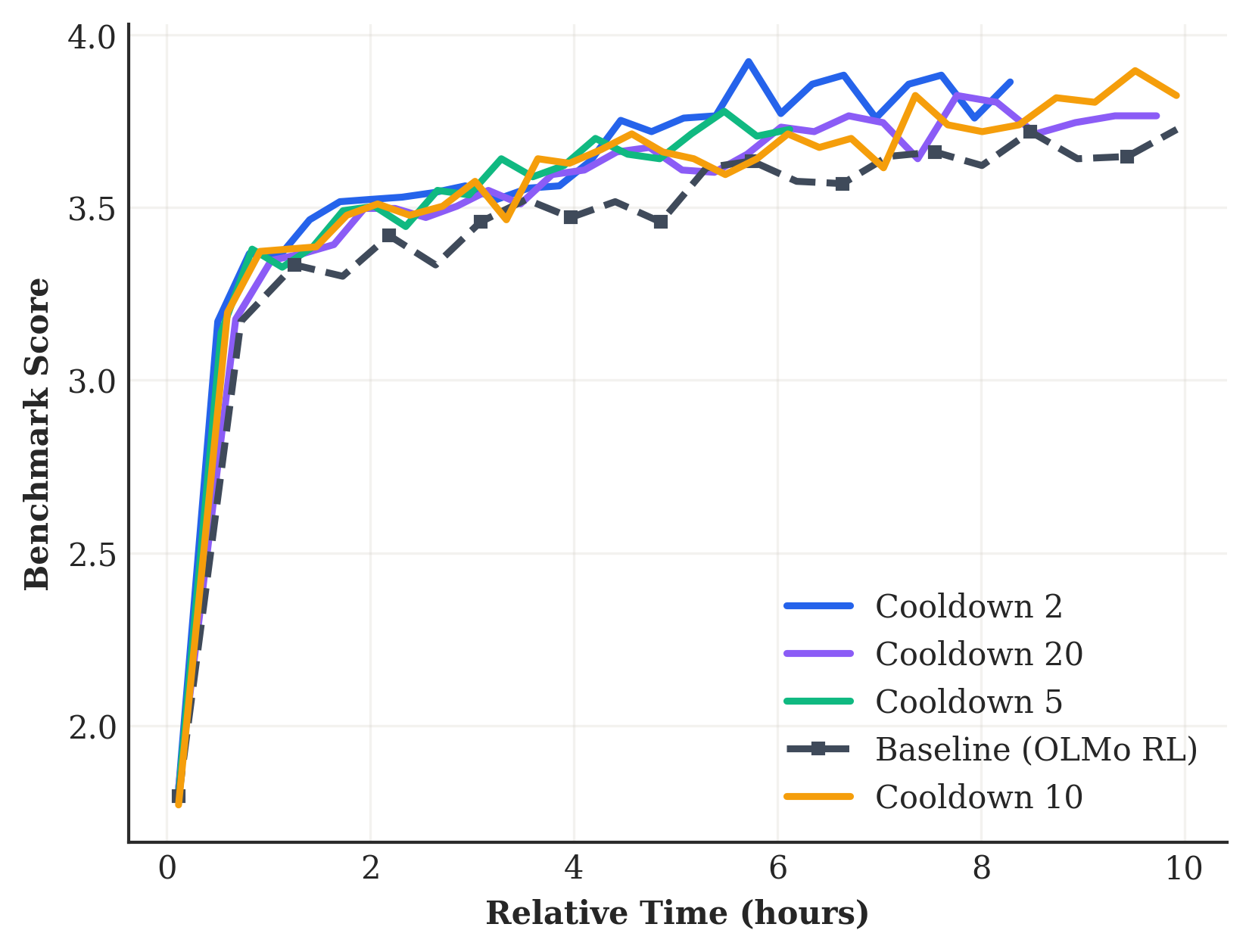}
        \caption{Different cooldown periods (2, 5, 10, 20) vs baseline. All cooldown periods are above the baseline, yet with marginal gains.}
        \label{fig:c_ablation_all}
    \end{subfigure}\hfill
    \begin{subfigure}[t]{0.45\linewidth}
        \centering
        \includegraphics[width=\linewidth]{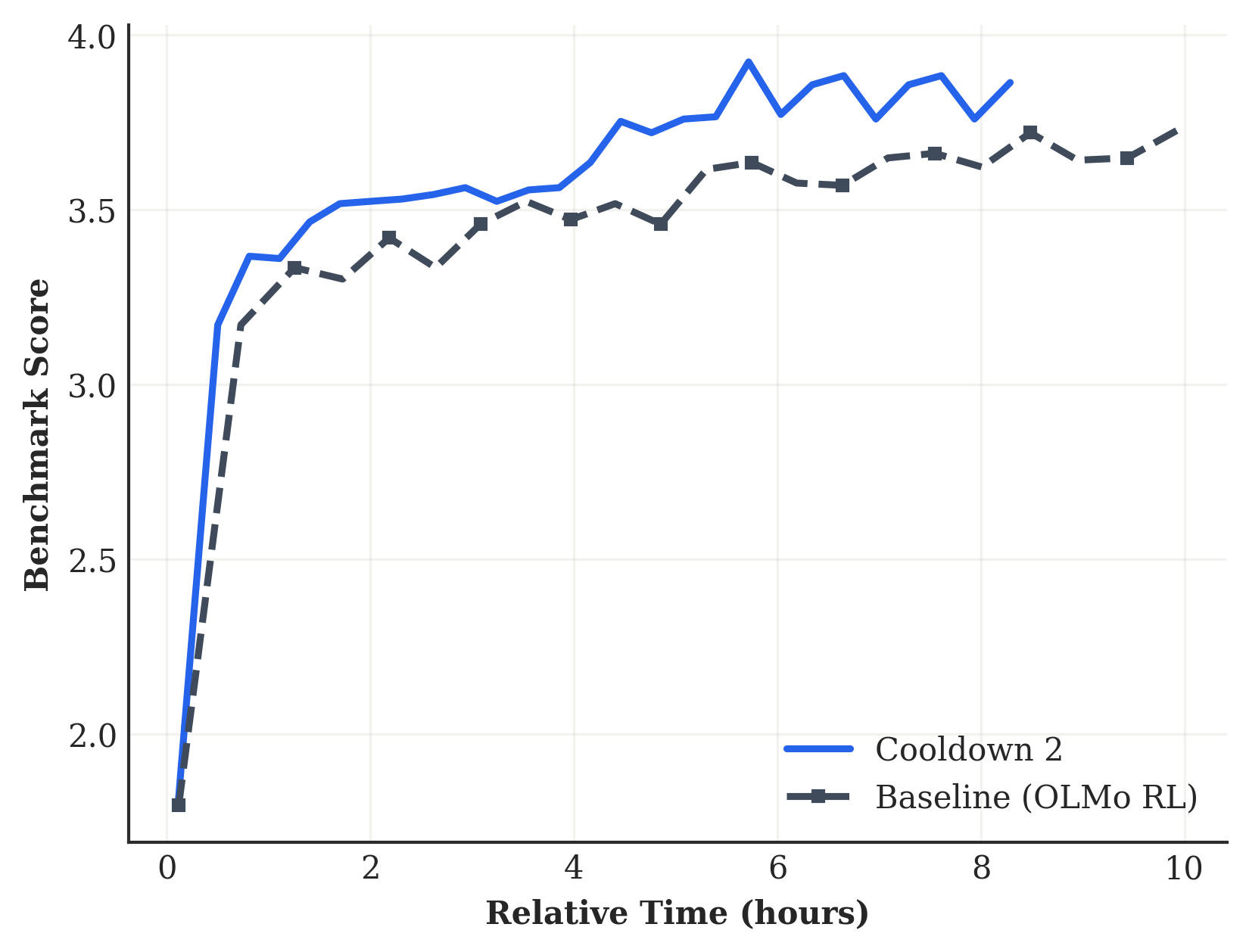}
        \caption{Only 2 cooldown steps prompt replay vs baseline, for readability. Surprisingly it does not show signs of overfitting and maintains better accuracy}
        \label{fig:c_ablation_cd2}
    \end{subfigure}

    \caption{Average accuracy over 6 benchmarks over time for cooldown-step ablation on Qwen2.5 1.5B, Dolci dataset; All cooldown steps have similar performance; The baseline was trained for 800 steps, a prompt was reused maximum 15 times.}
    \label{fig:c_ablation}
    \vspace{-0.1in}
\end{figure}

\subsection{Varying Cooldown Steps}

Our method comes with four hyperparameters. Cooldown period $C$ determines how often prompts re-occur, regulating overfitting. Fraction $\epsilon$ is the max prompts that can be reused in a batch. 
The pair $p_{\min},p_{\max}$ determines the range of pass-rates to filter prompts upon for the replay buffer and, finally, $R$ determines how often the algorithm is allowed to reuse certain prompts. 
We perform a sensitivity analysis only to the cooldown steps, using the values [2,5,10,20].

Figure \ref{fig:c_ablation} presents results of cooldown steps sensitivity analysis where different cooldown steps $C$ were tested [2, 5, 10, 20]. All cooldown steps have similar performance. This is a surprising result, as it was expected that a low value of 2 (meaning using the same prompt every other training step) would lead to overfitting. Yet, it performs as well as the other configurations. We investigate this further in the discussion.

\section{Discussion}

\subsection{Insensitivity to Cooldown steps}

\begin{wrapfigure}{r}{0.35\linewidth}
  \centering
  \vspace{-0.7in} 
  \includegraphics[width=\linewidth]{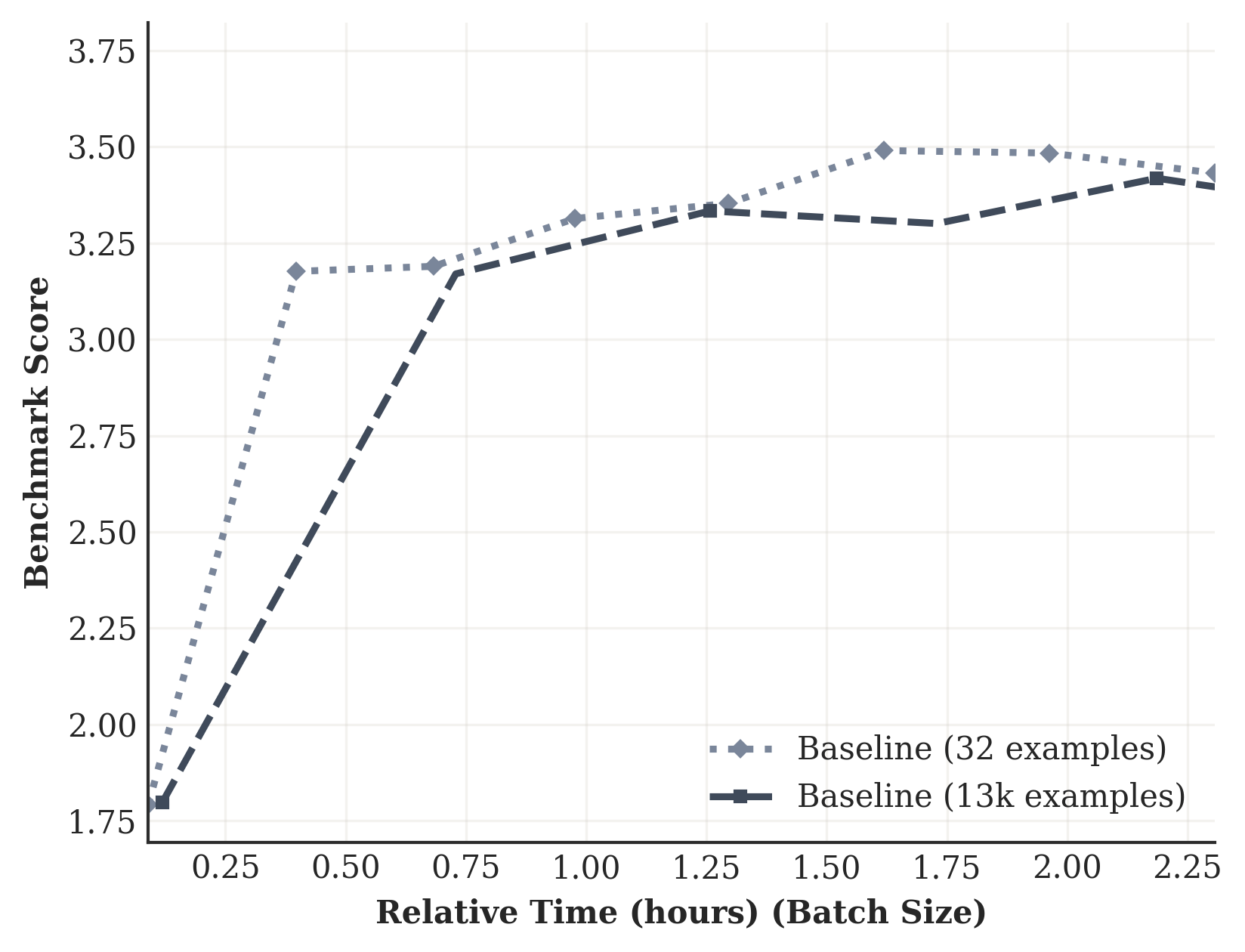}
  \caption{Avg Accuracy over 6 benchmarks, Baseline OLMo-RL Qwen 2.5 1.5B on Dolci, training on full dataset vs 32 prompts}
  \label{fig:32test}
  \vspace{-0.75\baselineskip} 
\end{wrapfigure}

Surprised by the results from Figure \ref{fig:c_ablation}, where 2 cooldown training steps were used, and overfitting did not appear, we decided to test the limits and try an experiment similar to 1 cooldown steps without a max reuse number - that is having the whole training size equal to the batch size. The same exact 32 prompts were used for training every single step, and compared with the baseline on the full dataset of 13k data points. Results are in Fig. \ref{fig:32test} and shockingly, it performs similarly or better than the baseline. This can be explained by spurious rewards: \citet{shao2025spuriousrewardsrethinkingtraining} explained the phenomenon, where Qwen 2.5 models, in particular, show an increase in performance even when giving random rewards; this does not happen with Llama and OLMo models; why this happens is not yet understood. Other papers try to take advantage of this phenomenon and come up with methods where a \textbf{single} data point (prompt) is being used and shows better accuracy \citep{li2026sampleruleallextreme,wang2025reinforcementlearningreasoninglarge}. 
For these reasons, we consider the ablations invalid and use different models for the main results. Small tests were done to ensure they are not susceptible to spurious rewards.

\subsection{Limitations \& Future Work}

Reflecting on the prompt replay method, although it is similar to the well-known Prioritized Experience Replay (PER), but without storing the actions (responses), only the states (prompts), it might be worth investigating what happens if, instead of using cooldown steps, the method would borrow more ideas from PER. For example, prompts could be stored in a buffer until it reaches a certain size, then each prompt would be given a probability of being reused based on its pass rate; this would make the sampling from the buffer stochastic, while the current prompt replay method uses a deterministic one.

Although sensitivity analysis was conducted on the cooldown steps, it proved unreliable. More robust hyperparameter optimization and ablations were not performed due to time and computational constraints. Future work could test the 4 hyperparameters mentioned in the method, and ablate prioritizing prompts with a pass rate closer to 0.5, to clearly separate the effect of speed-ups due to fewer zero-variance prompts and speed-ups due to higher advantage.

Another limitation is the number of model types, model sizes, and datasets. Even though we aimed at testing at least two of each, a more robust setting could have three model sizes (1B, 4B, 8B parameters), three model types (Llama, Qwen, OLMo), each with two datasets, as the method performance is highly dependent on the model's capabilities on a certain dataset, i.e. the number of prompts with pass rate 0.

Finally, a limiting factor in academia will always be computing power. Although we run the experiments until the learning curve flattened out, and most of the literature on GRPO and curriculum run experiments at a similar budget, works from \citet{olmo2025olmo3}  and \citet{khatri2025artscalingreinforcementlearning} on show that scaling the compute continues to bring more gains. The authors of the latter spent 100k GPU hours on a single 8B run, while our total budget for the project was 3k. Thus, we leave for future work to scale up the prompt replay method, or incorporate it with together with other methods in an industry-scale training run. 
Computing is also the reason multiple runs with different seeds were not performed to get statistical significance of the results, an unhealthy practice common in the literature. Yet, because the RL happens on an already trained model, \citet{khatri2025artscalingreinforcementlearning} claim the variance of GRPO variants is at most 2\% error margin.

\section{Conclusion}

We presented \textbf{Prompt Replay}, an overhead-free online data selection method for GRPO-style RLVR that reuses \emph{prompts only}. By reusing and prioritizing medium-difficulty prompts (pass rate near $0.5$), Prompt Replay reduces wasted rollouts on zero-variance prompts and increases the mean absolute advantage, yielding faster early gains in benchmark accuracy. We observe consistent improvements in rollout efficiency and initial learning speed, but aggressive reuse can lead to earlier plateauing and eventual convergence with the baseline. Finally, we find Qwen2.5-Math exhibits spurious-reward behavior that can invalidate ablations, suggesting it should not be used as the sole testbed for GRPO method research.

\subsection*{Acknowledgements}

We thank Thomas Moerland for valuable feedback and discussions that helped shape this work. We also thank LIACS and Rob van Nieuwpoort for granting access to Snellius, the Dutch national supercomputer, which made the experiments possible.

\bibliography{references}

\begin{thebibliography}{47}
\providecommand{\natexlab}[1]{#1}
\providecommand{\url}[1]{\texttt{#1}}
\expandafter\ifx\csname urlstyle\endcsname\relax
  \providecommand{\doi}[1]{doi: #1}\else
  \providecommand{\doi}{doi: \begingroup \urlstyle{rm}\Url}\fi

\bibitem[An et~al.()An, Xie, Li, Li, Zhang, Gong, Zhong, Xu, Qiu, Wang,
  et~al.]{an2025polaris}
Chenxin An, Zhihui Xie, Xiaonan Li, Lei Li, Jun Zhang, Shansan Gong, Ming
  Zhong, Jingjing Xu, Xipeng Qiu, Mingxuan Wang, et~al.
\newblock Polaris: A post-training recipe for scaling reinforcement learning on
  advanced reasoning models, 2025.
\newblock \emph{URL https://hkunlp.github.io/blog/2025/Polaris}.

\bibitem[An et~al.(2025)An, Xie, Li, Li, Zhang, Gong, Zhong, Xu, Qiu, Wang, and
  Kong]{Polaris2025}
Chenxin An, Zhihui Xie, Xiaonan Li, Lei Li, Jun Zhang, Shansan Gong, Ming
  Zhong, Jingjing Xu, Xipeng Qiu, Mingxuan Wang, and Lingpeng Kong.
\newblock Polaris: A post-training recipe for scaling reinforcement learning on
  advanced reasoning models, 2025.
\newblock URL \url{https://hkunlp.github.io/blog/2025/Polaris}.

\bibitem[Bae et~al.(2025)Bae, Hong, Lee, Kim, Nam, and Kwak]{bae2025online}
Sanghwan Bae, Jiwoo Hong, Min~Young Lee, Hanbyul Kim, JeongYeon Nam, and
  Donghyun Kwak.
\newblock Online difficulty filtering for reasoning oriented reinforcement
  learning.
\newblock \emph{arXiv preprint arXiv:2504.03380}, 2025.

\bibitem[Bai et~al.(2022)Bai, Jones, Ndousse, Askell, Chen, DasSarma, Drain,
  Fort, Ganguli, Henighan, Joseph, Kadavath, Kernion, Conerly, El-Showk,
  Elhage, Hatfield-Dodds, Hernandez, Hume, Johnston, Kravec, Lovitt, Nanda,
  Olsson, Amodei, Brown, Clark, McCandlish, Olah, Mann, and
  Kaplan]{bai2022traininghelpfulharmlessassistant}
Yuntao Bai, Andy Jones, Kamal Ndousse, Amanda Askell, Anna Chen, Nova DasSarma,
  Dawn Drain, Stanislav Fort, Deep Ganguli, Tom Henighan, Nicholas Joseph,
  Saurav Kadavath, Jackson Kernion, Tom Conerly, Sheer El-Showk, Nelson Elhage,
  Zac Hatfield-Dodds, Danny Hernandez, Tristan Hume, Scott Johnston, Shauna
  Kravec, Liane Lovitt, Neel Nanda, Catherine Olsson, Dario Amodei, Tom Brown,
  Jack Clark, Sam McCandlish, Chris Olah, Ben Mann, and Jared Kaplan.
\newblock Training a helpful and harmless assistant with reinforcement learning
  from human feedback, 2022.
\newblock URL \url{https://arxiv.org/abs/2204.05862}.

\bibitem[Bengio et~al.(2009)Bengio, Louradour, Collobert, and
  Weston]{bengio2009curriculum}
Yoshua Bengio, J{\'e}r{\^o}me Louradour, Ronan Collobert, and Jason Weston.
\newblock Curriculum learning.
\newblock In \emph{Proceedings of the 26th annual international conference on
  machine learning}, pp.\  41--48, 2009.

\bibitem[Christiano et~al.(2023)Christiano, Leike, Brown, Martic, Legg, and
  Amodei]{christiano2023deepreinforcementlearninghuman}
Paul Christiano, Jan Leike, Tom~B. Brown, Miljan Martic, Shane Legg, and Dario
  Amodei.
\newblock Deep reinforcement learning from human preferences, 2023.
\newblock URL \url{https://arxiv.org/abs/1706.03741}.

\bibitem[Dou et~al.(2025)Dou, Wu, Xu, Zheng, Gui, Zhang, and
  Huang]{dou2025improvingrlexplorationllm}
Shihan Dou, Muling Wu, Jingwen Xu, Rui Zheng, Tao Gui, Qi~Zhang, and Xuanjing
  Huang.
\newblock Improving rl exploration for llm reasoning through retrospective
  replay, 2025.
\newblock URL \url{https://arxiv.org/abs/2504.14363}.

\bibitem[Foster et~al.(2025)Foster, Sims, Forkel, Fellows, and
  Foerster]{foster2025learningreasonfrontierlearnability}
Thomas Foster, Anya Sims, Johannes Forkel, Mattie Fellows, and Jakob Foerster.
\newblock Learning to reason at the frontier of learnability, 2025.
\newblock URL \url{https://arxiv.org/abs/2502.12272}.

\bibitem[Gao et~al.(2025)Gao, Kim, Sun, Joachims, Wang, Pang, and
  Tan]{gao2025prompt}
Zhaolin Gao, Joongwon Kim, Wen Sun, Thorsten Joachims, Sid Wang,
  Richard~Yuanzhe Pang, and Liang Tan.
\newblock Prompt curriculum learning for efficient llm post-training.
\newblock \emph{arXiv preprint arXiv:2510.01135}, 2025.

\bibitem[Guo et~al.(2025)Guo, Yang, Zhang, Song, Wang, Zhu, Xu, Zhang, Ma, Bi,
  Zhang, Yu, Wu, Wu, Gou, Shao, Li, Gao, Liu, Xue, Wang, Wu, Feng, Lu, Zhao,
  Deng, Ruan, Dai, Chen, Ji, Li, Lin, Dai, Luo, Hao, Chen, Li, Zhang, Xu, Ding,
  Gao, Qu, Li, Guo, Li, Chen, Yuan, Tu, Qiu, Li, Cai, Ni, Liang, Chen, Dong,
  Hu, You, Gao, Guan, Huang, Yu, Wang, Zhang, Zhao, Wang, Zhang, Xu, Xia,
  Zhang, Zhang, Tang, Zhou, Li, Wang, Li, Tian, Huang, Zhang, Wang, Chen, Du,
  Ge, Zhang, Pan, Wang, Chen, Jin, Chen, Lu, Zhou, Chen, Ye, Wang, Yu, Zhou,
  Pan, Li, Zhou, Wu, Yun, Pei, Sun, Wang, Zeng, Liu, Liang, Gao, Yu, Zhang,
  Xiao, An, Liu, Wang, Chen, Nie, Cheng, Liu, Xie, Liu, Yang, Li, Su, Lin, Li,
  Jin, Shen, Chen, Sun, Wang, Song, Zhou, Wang, Shan, Li, Wang, Wei, Zhang, Xu,
  Li, Zhao, Sun, Wang, Yu, Zhang, Shi, Xiong, He, Piao, Wang, Tan, Ma, Liu,
  Guo, Ou, Wang, Gong, Zou, He, Xiong, Luo, You, Liu, Zhou, Zhu, Huang, Li,
  Zheng, Zhu, Ma, Tang, Zha, Yan, Ren, Ren, Sha, Fu, Xu, Xie, Zhang, Hao, Ma,
  Yan, Wu, Gu, Zhu, Liu, Li, Xie, Song, Pan, Huang, Xu, Zhang, and
  Zhang]{Guo_2025}
Daya Guo, Dejian Yang, Haowei Zhang, Junxiao Song, Peiyi Wang, Qihao Zhu,
  Runxin Xu, Ruoyu Zhang, Shirong Ma, Xiao Bi, Xiaokang Zhang, Xingkai Yu,
  Yu~Wu, Z.~F. Wu, Zhibin Gou, Zhihong Shao, Zhuoshu Li, Ziyi Gao, Aixin Liu,
  Bing Xue, Bingxuan Wang, Bochao Wu, Bei Feng, Chengda Lu, Chenggang Zhao,
  Chengqi Deng, Chong Ruan, Damai Dai, Deli Chen, Dongjie Ji, Erhang Li,
  Fangyun Lin, Fucong Dai, Fuli Luo, Guangbo Hao, Guanting Chen, Guowei Li,
  H.~Zhang, Hanwei Xu, Honghui Ding, Huazuo Gao, Hui Qu, Hui Li, Jianzhong Guo,
  Jiashi Li, Jingchang Chen, Jingyang Yuan, Jinhao Tu, Junjie Qiu, Junlong Li,
  J.~L. Cai, Jiaqi Ni, Jian Liang, Jin Chen, Kai Dong, Kai Hu, Kaichao You,
  Kaige Gao, Kang Guan, Kexin Huang, Kuai Yu, Lean Wang, Lecong Zhang, Liang
  Zhao, Litong Wang, Liyue Zhang, Lei Xu, Leyi Xia, Mingchuan Zhang, Minghua
  Zhang, Minghui Tang, Mingxu Zhou, Meng Li, Miaojun Wang, Mingming Li, Ning
  Tian, Panpan Huang, Peng Zhang, Qiancheng Wang, Qinyu Chen, Qiushi Du, Ruiqi
  Ge, Ruisong Zhang, Ruizhe Pan, Runji Wang, R.~J. Chen, R.~L. Jin, Ruyi Chen,
  Shanghao Lu, Shangyan Zhou, Shanhuang Chen, Shengfeng Ye, Shiyu Wang,
  Shuiping Yu, Shunfeng Zhou, Shuting Pan, S.~S. Li, Shuang Zhou, Shaoqing Wu,
  Tao Yun, Tian Pei, Tianyu Sun, T.~Wang, Wangding Zeng, Wen Liu, Wenfeng
  Liang, Wenjun Gao, Wenqin Yu, Wentao Zhang, W.~L. Xiao, Wei An, Xiaodong Liu,
  Xiaohan Wang, Xiaokang Chen, Xiaotao Nie, Xin Cheng, Xin Liu, Xin Xie,
  Xingchao Liu, Xinyu Yang, Xinyuan Li, Xuecheng Su, Xuheng Lin, X.~Q. Li,
  Xiangyue Jin, Xiaojin Shen, Xiaosha Chen, Xiaowen Sun, Xiaoxiang Wang, Xinnan
  Song, Xinyi Zhou, Xianzu Wang, Xinxia Shan, Y.~K. Li, Y.~Q. Wang, Y.~X. Wei,
  Yang Zhang, Yanhong Xu, Yao Li, Yao Zhao, Yaofeng Sun, Yaohui Wang, Yi~Yu,
  Yichao Zhang, Yifan Shi, Yiliang Xiong, Ying He, Yishi Piao, Yisong Wang,
  Yixuan Tan, Yiyang Ma, Yiyuan Liu, Yongqiang Guo, Yuan Ou, Yuduan Wang, Yue
  Gong, Yuheng Zou, Yujia He, Yunfan Xiong, Yuxiang Luo, Yuxiang You, Yuxuan
  Liu, Yuyang Zhou, Y.~X. Zhu, Yanping Huang, Yaohui Li, Yi~Zheng, Yuchen Zhu,
  Yunxian Ma, Ying Tang, Yukun Zha, Yuting Yan, Z.~Z. Ren, Zehui Ren, Zhangli
  Sha, Zhe Fu, Zhean Xu, Zhenda Xie, Zhengyan Zhang, Zhewen Hao, Zhicheng Ma,
  Zhigang Yan, Zhiyu Wu, Zihui Gu, Zijia Zhu, Zijun Liu, Zilin Li, Ziwei Xie,
  Ziyang Song, Zizheng Pan, Zhen Huang, Zhipeng Xu, Zhongyu Zhang, and Zhen
  Zhang.
\newblock Deepseek-r1 incentivizes reasoning in llms through reinforcement
  learning.
\newblock \emph{Nature}, 645\penalty0 (8081):\penalty0 633–638, September
  2025.
\newblock ISSN 1476-4687.
\newblock \doi{10.1038/s41586-025-09422-z}.
\newblock URL \url{http://dx.doi.org/10.1038/s41586-025-09422-z}.

\bibitem[He et~al.(2024)He, Luo, Bai, Hu, Thai, Shen, Hu, Han, Huang, Zhang,
  Liu, Qi, Liu, and Sun]{olympiadbench}
Chaoqun He, Renjie Luo, Yuzhuo Bai, Shengding Hu, Zhen Thai, Junhao Shen, Jinyi
  Hu, Xu~Han, Yujie Huang, Yuxiang Zhang, Jie Liu, Lei Qi, Zhiyuan Liu, and
  Maosong Sun.
\newblock {O}lympiad{B}ench: A challenging benchmark for promoting {AGI} with
  olympiad-level bilingual multimodal scientific problems.
\newblock In \emph{Proceedings of the 62nd Annual Meeting of the Association
  for Computational Linguistics (Volume 1: Long Papers)}, pp.\  3828--3850,
  Bangkok, Thailand, August 2024. Association for Computational Linguistics.
\newblock \doi{10.18653/v1/2024.acl-long.211}.
\newblock URL \url{https://aclanthology.org/2024.acl-long.211/}.

\bibitem[Khatri et~al.(2025)Khatri, Madaan, Tiwari, Bansal, Duvvuri, Zaheer,
  Dhillon, Brandfonbrener, and
  Agarwal]{khatri2025artscalingreinforcementlearning}
Devvrit Khatri, Lovish Madaan, Rishabh Tiwari, Rachit Bansal, Sai~Surya
  Duvvuri, Manzil Zaheer, Inderjit~S. Dhillon, David Brandfonbrener, and
  Rishabh Agarwal.
\newblock The art of scaling reinforcement learning compute for llms, 2025.
\newblock URL \url{https://arxiv.org/abs/2510.13786}.

\bibitem[Lambert et~al.(2025)Lambert, Morrison, Pyatkin, Huang, Ivison,
  Brahman, Miranda, Liu, Dziri, Lyu, Gu, Malik, Graf, Hwang, Yang, Bras,
  Tafjord, Wilhelm, Soldaini, Smith, Wang, Dasigi, and
  Hajishirzi]{lambert2025tulu3pushingfrontiers}
Nathan Lambert, Jacob Morrison, Valentina Pyatkin, Shengyi Huang, Hamish
  Ivison, Faeze Brahman, Lester James~V. Miranda, Alisa Liu, Nouha Dziri, Shane
  Lyu, Yuling Gu, Saumya Malik, Victoria Graf, Jena~D. Hwang, Jiangjiang Yang,
  Ronan~Le Bras, Oyvind Tafjord, Chris Wilhelm, Luca Soldaini, Noah~A. Smith,
  Yizhong Wang, Pradeep Dasigi, and Hannaneh Hajishirzi.
\newblock Tulu 3: Pushing frontiers in open language model post-training, 2025.
\newblock URL \url{https://arxiv.org/abs/2411.15124}.

\bibitem[Lewkowycz et~al.(2022)Lewkowycz, Andreassen, Dohan, Dyer, Michalewski,
  Ramasesh, Slone, Anil, Schlag, Gutman-Solo, Wu, Neyshabur, Gur-Ari, and
  Misra]{minervamath}
Aitor Lewkowycz, Anders Andreassen, David Dohan, Ethan Dyer, Henryk
  Michalewski, Vinay~V. Ramasesh, Ambrose Slone, Cem Anil, Imanol Schlag, Theo
  Gutman-Solo, Yuhuai Wu, Behnam Neyshabur, Guy Gur-Ari, and Vedant Misra.
\newblock Solving quantitative reasoning problems with language models.
\newblock In \emph{Advances in Neural Information Processing Systems 35
  (NeurIPS 2022)}, 2022.
\newblock URL
  \url{http://papers.nips.cc/paper_files/paper/2022/hash/18abbeef8cfe9203fdf9053c9c4fe191-Abstract-Conference.html}.

\bibitem[Li et~al.(2025)Li, Zou, and Liu]{li2025limr}
Xuefeng Li, Haoyang Zou, and Pengfei Liu.
\newblock Limr: Less is more for rl scaling.
\newblock \emph{arXiv preprint arXiv:2502.11886}, 2025.

\bibitem[Li et~al.(2026)Li, Huang, Wu, Wang, Li, Luo, Su, Zheng, and
  Liu]{li2026sampleruleallextreme}
Yiyuan Li, Zhen Huang, Yanan Wu, Weixun Wang, Xuefeng Li, Yijia Luo, Wenbo Su,
  Bo~Zheng, and Pengfei Liu.
\newblock One sample to rule them all: Extreme data efficiency in rl scaling,
  2026.
\newblock URL \url{https://arxiv.org/abs/2601.03111}.

\bibitem[Lightman et~al.(2024)Lightman, Kosaraju, Burda, Edwards, Baker, Lee,
  Leike, Schulman, Sutskever, and Cobbe]{math500}
Hunter Lightman, Vineet Kosaraju, Yura Burda, Harrison Edwards, Bowen Baker,
  Teddy Lee, Jan Leike, John Schulman, Ilya Sutskever, and Karl Cobbe.
\newblock Let’s verify step by step.
\newblock In \emph{The Twelfth International Conference on Learning
  Representations (ICLR)}, 2024.
\newblock URL \url{https://openreview.net/forum?id=v8L0pN6EOi}.
\newblock Introduces PRM800K and the MATH-500 evaluation split.

\bibitem[Lin(1992)]{lin1992self}
Long-Ji Lin.
\newblock Self-improving reactive agents based on reinforcement learning,
  planning and teaching.
\newblock \emph{Machine learning}, 8\penalty0 (3):\penalty0 293--321, 1992.

\bibitem[Liu et~al.(2025{\natexlab{a}})Liu, Diao, Lu, Hu, Dong, Choi, Kautz,
  and Dong]{liu2025prorl}
Mingjie Liu, Shizhe Diao, Ximing Lu, Jian Hu, Xin Dong, Yejin Choi, Jan Kautz,
  and Yi~Dong.
\newblock Prorl: Prolonged reinforcement learning expands reasoning boundaries
  in large language models.
\newblock \emph{arXiv preprint arXiv:2505.24864}, 2025{\natexlab{a}}.

\bibitem[Liu et~al.(2025{\natexlab{b}})Liu, Chen, Li, Qi, Pang, Du, Lee, and
  Lin]{liu2025understandingr1zeroliketrainingcritical}
Zichen Liu, Changyu Chen, Wenjun Li, Penghui Qi, Tianyu Pang, Chao Du, Wee~Sun
  Lee, and Min Lin.
\newblock Understanding r1-zero-like training: A critical perspective,
  2025{\natexlab{b}}.
\newblock URL \url{https://arxiv.org/abs/2503.20783}.

\bibitem[Luo et~al.(2025)Luo, Tan, Wong, Shi, Tang, Roongta, Cai, Luo, Zhang,
  Li, et~al.]{luo2025deepscaler}
Michael Luo, Sijun Tan, Justin Wong, Xiaoxiang Shi, William~Y Tang, Manan
  Roongta, Colin Cai, Jeffrey Luo, Tianjun Zhang, Li~Erran Li, et~al.
\newblock Deepscaler: Surpassing o1-preview with a 1.5 b model by scaling rl.
\newblock \emph{Notion Blog}, 2025.

\bibitem[{Meta}(2024)]{llama32_3b_instruct_hf}
{Meta}.
\newblock meta-llama/llama-3.2-3b-instruct (model card).
\newblock \url{https://huggingface.co/meta-llama/Llama-3.2-3B-Instruct}, 2024.
\newblock Accessed 2026-01-29.

\bibitem[Muennighoff et~al.(2025)Muennighoff, Yang, Shi, Li, Fei-Fei,
  Hajishirzi, Zettlemoyer, Liang, Candès, and
  Hashimoto]{muennighoff2025s1simpletesttimescaling}
Niklas Muennighoff, Zitong Yang, Weijia Shi, Xiang~Lisa Li, Li~Fei-Fei,
  Hannaneh Hajishirzi, Luke Zettlemoyer, Percy Liang, Emmanuel Candès, and
  Tatsunori Hashimoto.
\newblock s1: Simple test-time scaling, 2025.
\newblock URL \url{https://arxiv.org/abs/2501.19393}.

\bibitem[Olmo et~al.(2025)Olmo, :, Ettinger, Bertsch, Kuehl, Graham, Heineman,
  Groeneveld, Brahman, Timbers, Ivison, Morrison, Poznanski, Lo, Soldaini,
  Jordan, Chen, Noukhovitch, Lambert, Walsh, Dasigi, Berry, Malik, Shah, Geng,
  Arora, Gupta, Anderson, Xiao, Murray, Romero, Graf, Asai, Bhagia, Wettig,
  Liu, Rangapur, Anastasiades, Huang, Schwenk, Trivedi, Magnusson, Lochner,
  Liu, Miranda, Sap, Morgan, Schmitz, Guerquin, Wilson, Huff, Bras, Xin, Shao,
  Skjonsberg, Shen, Li, Wilde, Pyatkin, Merrill, Chang, Gu, Zeng, Sabharwal,
  Zettlemoyer, Koh, Farhadi, Smith, and Hajishirzi]{olmo2025olmo3}
Team Olmo, :, Allyson Ettinger, Amanda Bertsch, Bailey Kuehl, David Graham,
  David Heineman, Dirk Groeneveld, Faeze Brahman, Finbarr Timbers, Hamish
  Ivison, Jacob Morrison, Jake Poznanski, Kyle Lo, Luca Soldaini, Matt Jordan,
  Mayee Chen, Michael Noukhovitch, Nathan Lambert, Pete Walsh, Pradeep Dasigi,
  Robert Berry, Saumya Malik, Saurabh Shah, Scott Geng, Shane Arora, Shashank
  Gupta, Taira Anderson, Teng Xiao, Tyler Murray, Tyler Romero, Victoria Graf,
  Akari Asai, Akshita Bhagia, Alexander Wettig, Alisa Liu, Aman Rangapur, Chloe
  Anastasiades, Costa Huang, Dustin Schwenk, Harsh Trivedi, Ian Magnusson,
  Jaron Lochner, Jiacheng Liu, Lester James~V. Miranda, Maarten Sap, Malia
  Morgan, Michael Schmitz, Michal Guerquin, Michael Wilson, Regan Huff,
  Ronan~Le Bras, Rui Xin, Rulin Shao, Sam Skjonsberg, Shannon~Zejiang Shen,
  Shuyue~Stella Li, Tucker Wilde, Valentina Pyatkin, Will Merrill, Yapei Chang,
  Yuling Gu, Zhiyuan Zeng, Ashish Sabharwal, Luke Zettlemoyer, Pang~Wei Koh,
  Ali Farhadi, Noah~A. Smith, and Hannaneh Hajishirzi.
\newblock Olmo 3, 2025.
\newblock URL \url{https://arxiv.org/abs/2512.13961}.

\bibitem[Ouyang et~al.(2022)Ouyang, Wu, Jiang, Almeida, Wainwright, Mishkin,
  Zhang, Agarwal, Slama, Ray, Schulman, Hilton, Kelton, Miller, Simens, Askell,
  Welinder, Christiano, Leike, and
  Lowe]{ouyang2022traininglanguagemodelsfollow}
Long Ouyang, Jeff Wu, Xu~Jiang, Diogo Almeida, Carroll~L. Wainwright, Pamela
  Mishkin, Chong Zhang, Sandhini Agarwal, Katarina Slama, Alex Ray, John
  Schulman, Jacob Hilton, Fraser Kelton, Luke Miller, Maddie Simens, Amanda
  Askell, Peter Welinder, Paul Christiano, Jan Leike, and Ryan Lowe.
\newblock Training language models to follow instructions with human feedback,
  2022.
\newblock URL \url{https://arxiv.org/abs/2203.02155}.

\bibitem[Qu et~al.(2025)Qu, Wang, Mao, Hu, Ommer, and Ji]{qu2025can}
Yun Qu, Qi~Wang, Yixiu Mao, Vincent~Tao Hu, Bj{\"o}rn Ommer, and Xiangyang Ji.
\newblock Can prompt difficulty be online predicted for accelerating rl
  finetuning of reasoning models?
\newblock \emph{arXiv preprint arXiv:2507.04632}, 2025.

\bibitem[Schaul et~al.(2016)Schaul, Quan, Antonoglou, and
  Silver]{schaul2016prioritizedexperiencereplay}
Tom Schaul, John Quan, Ioannis Antonoglou, and David Silver.
\newblock Prioritized experience replay, 2016.
\newblock URL \url{https://arxiv.org/abs/1511.05952}.

\bibitem[Shao et~al.(2025)Shao, Li, Xin, Geng, Wang, Oh, Du, Lambert, Min,
  Krishna, Tsvetkov, Hajishirzi, Koh, and
  Zettlemoyer]{shao2025spuriousrewardsrethinkingtraining}
Rulin Shao, Shuyue~Stella Li, Rui Xin, Scott Geng, Yiping Wang, Sewoong Oh,
  Simon~Shaolei Du, Nathan Lambert, Sewon Min, Ranjay Krishna, Yulia Tsvetkov,
  Hannaneh Hajishirzi, Pang~Wei Koh, and Luke Zettlemoyer.
\newblock Spurious rewards: Rethinking training signals in rlvr, 2025.
\newblock URL \url{https://arxiv.org/abs/2506.10947}.

\bibitem[Shao et~al.(2024)Shao, Wang, Zhu, Xu, Song, Bi, Zhang, Zhang, Li, Wu,
  and Guo]{shao2024deepseekmathpushinglimitsmathematical}
Zhihong Shao, Peiyi Wang, Qihao Zhu, Runxin Xu, Junxiao Song, Xiao Bi, Haowei
  Zhang, Mingchuan Zhang, Y.~K. Li, Y.~Wu, and Daya Guo.
\newblock Deepseekmath: Pushing the limits of mathematical reasoning in open
  language models, 2024.
\newblock URL \url{https://arxiv.org/abs/2402.03300}.

\bibitem[Shi et~al.(2025)Shi, Wu, Song, Zhou, and Zhao]{shi2025efficient}
Taiwei Shi, Yiyang Wu, Linxin Song, Tianyi Zhou, and Jieyu Zhao.
\newblock Efficient reinforcement finetuning via adaptive curriculum learning.
\newblock \emph{arXiv preprint arXiv:2504.05520}, 2025.

\bibitem[Song et~al.(2025)Song, Zheng, Li, Yang, Luo, Pan, and
  Zhang]{song2025fastcurlcurriculumreinforcementlearning}
Mingyang Song, Mao Zheng, Zheng Li, Wenjie Yang, Xuan Luo, Yue Pan, and Feng
  Zhang.
\newblock Fastcurl: Curriculum reinforcement learning with stage-wise context
  scaling for efficient training r1-like reasoning models, 2025.
\newblock URL \url{https://arxiv.org/abs/2503.17287}.

\bibitem[Sun et~al.(2025)Sun, Shen, Wang, Chen, Wang, Zhou, and
  Zhang]{sun2025improving}
Yifan Sun, Jingyan Shen, Yibin Wang, Tianyu Chen, Zhendong Wang, Mingyuan Zhou,
  and Huan Zhang.
\newblock Improving data efficiency for llm reinforcement fine-tuning through
  difficulty-targeted online data selection and rollout replay.
\newblock \emph{arXiv preprint arXiv:2506.05316}, 2025.

\bibitem[Team et~al.(2025)Team, Du, Gao, Xing, Jiang, Chen, Li, Xiao, Du, Liao,
  et~al.]{team2025kimi}
Kimi Team, Angang Du, Bofei Gao, Bowei Xing, Changjiu Jiang, Cheng Chen, Cheng
  Li, Chenjun Xiao, Chenzhuang Du, Chonghua Liao, et~al.
\newblock Kimi k1. 5: Scaling reinforcement learning with llms.
\newblock \emph{arXiv preprint arXiv:2501.12599}, 2025.

\bibitem[Wang et~al.(2025)Wang, Yang, Zeng, Ren, Liu, Peng, Cheng, He, Wang,
  Gao, Chen, Wang, Du, and Shen]{wang2025reinforcementlearningreasoninglarge}
Yiping Wang, Qing Yang, Zhiyuan Zeng, Liliang Ren, Liyuan Liu, Baolin Peng, Hao
  Cheng, Xuehai He, Kuan Wang, Jianfeng Gao, Weizhu Chen, Shuohang Wang,
  Simon~Shaolei Du, and Yelong Shen.
\newblock Reinforcement learning for reasoning in large language models with
  one training example, 2025.
\newblock URL \url{https://arxiv.org/abs/2504.20571}.

\bibitem[Yang et~al.(2024)Yang, Zhang, Hui, Gao, Yu, Li, Liu, Tu, Zhou, Lin,
  Lu, Xue, Lin, Liu, Ren, and Zhang]{yang2024qwen25math}
An~Yang, Beichen Zhang, Binyuan Hui, Bofei Gao, Bowen Yu, Chengpeng Li,
  Dayiheng Liu, Jianhong Tu, Jingren Zhou, Junyang Lin, Keming Lu, Mingfeng
  Xue, Runji Lin, Tianyu Liu, Xingzhang Ren, and Zhenru Zhang.
\newblock Qwen2.5-math technical report: Toward mathematical expert model via
  self-improvement.
\newblock 2024.

\bibitem[Yang et~al.(2025)]{yang2025qwen3}
An~Yang et~al.
\newblock Qwen3 technical report.
\newblock 2025.

\bibitem[Yao et~al.(2025)Yao, Liu, Zhang, Dong, Shang, and Gao]{yao2025your}
Feng Yao, Liyuan Liu, Dinghuai Zhang, Chengyu Dong, Jingbo Shang, and Jianfeng
  Gao.
\newblock Your efficient rl framework secretly brings you off-policy rl
  training.
\newblock \emph{Feng Yao’s Notion}, 2025.

\bibitem[Yu et~al.(2025)Yu, Zhang, Zhu, Yuan, Zuo, Yue, Dai, Fan, Liu, Liu,
  et~al.]{yu2025dapo}
Qiying Yu, Zheng Zhang, Ruofei Zhu, Yufeng Yuan, Xiaochen Zuo, Yu~Yue, Weinan
  Dai, Tiantian Fan, Gaohong Liu, Lingjun Liu, et~al.
\newblock Dapo: An open-source llm reinforcement learning system at scale.
\newblock \emph{arXiv preprint arXiv:2503.14476}, 2025.

\bibitem[Zeng et~al.(2025)Zeng, Sun, Ji, Min, Cai, Wang, Yin, Zhang, Chen, and
  Wang]{zeng2025cures}
Yongcheng Zeng, Zexu Sun, Bokai Ji, Erxue Min, Hengyi Cai, Shuaiqiang Wang,
  Dawei Yin, Haifeng Zhang, Xu~Chen, and Jun Wang.
\newblock Cures: From gradient analysis to efficient curriculum learning for
  reasoning llms.
\newblock \emph{arXiv preprint arXiv:2510.01037}, 2025.

\bibitem[Zhan et~al.(2025)Zhan, Li, Wang, Qu, Liu, Shao, Wong, and
  Cheng]{zhan2025exgrpolearningreasonexperience}
Runzhe Zhan, Yafu Li, Zhi Wang, Xiaoye Qu, Dongrui Liu, Jing Shao, Derek~F.
  Wong, and Yu~Cheng.
\newblock Exgrpo: Learning to reason from experience, 2025.
\newblock URL \url{https://arxiv.org/abs/2510.02245}.

\bibitem[Zhang et~al.(2025{\natexlab{a}})Zhang, Fu, Zhang, Fu, Wang, Zhang, and
  Zhou]{zhang2025rlepreinforcementlearningexperience}
Hongzhi Zhang, Jia Fu, Jingyuan Zhang, Kai Fu, Qi~Wang, Fuzheng Zhang, and
  Guorui Zhou.
\newblock Rlep: Reinforcement learning with experience replay for llm
  reasoning, 2025{\natexlab{a}}.
\newblock URL \url{https://arxiv.org/abs/2507.07451}.

\bibitem[Zhang et~al.(2025{\natexlab{b}})Zhang, Arora, Mei, and
  Zanette]{zhang2025speed}
Ruiqi Zhang, Daman Arora, Song Mei, and Andrea Zanette.
\newblock Speed-rl: Faster training of reasoning models via online curriculum
  learning.
\newblock \emph{arXiv preprint arXiv:2506.09016}, 2025{\natexlab{b}}.

\bibitem[Zhang \& Math-AI(2023)Zhang and Math-AI]{amc23}
Yifan Zhang and Team Math-AI.
\newblock American mathematics competition (amc) 2023.
\newblock \url{https://huggingface.co/datasets/math-ai/amc23}, 2023.
\newblock Accessed 2026-01-28.

\bibitem[Zhang \& Math-AI(2024)Zhang and Math-AI]{aime24}
Yifan Zhang and Team Math-AI.
\newblock American invitational mathematics examination (aime) 2024.
\newblock \url{https://huggingface.co/datasets/math-ai/aime24}, 2024.
\newblock Accessed 2026-01-28.

\bibitem[Zhang \& Math-AI(2025)Zhang and Math-AI]{aime25}
Yifan Zhang and Team Math-AI.
\newblock American invitational mathematics examination (aime) 2025.
\newblock \url{https://huggingface.co/datasets/math-ai/aime25}, 2025.
\newblock Accessed 2026-01-28.

\bibitem[Zhang et~al.(2025{\natexlab{c}})Zhang, Yao, Yu, Liu, Yin, Yin, Yun,
  and Li]{zhang2025improvingsamplingefficiencyrlvr}
Yuheng Zhang, Wenlin Yao, Changlong Yu, Yao Liu, Qingyu Yin, Bing Yin, Hyokun
  Yun, and Lihong Li.
\newblock Improving sampling efficiency in rlvr through adaptive rollout and
  response reuse, 2025{\natexlab{c}}.
\newblock URL \url{https://arxiv.org/abs/2509.25808}.

\bibitem[Zheng et~al.(2025)Zheng, Zhou, Bartoldson, Kailkhura, Lai, Zhao, and
  Chen]{zheng2025act}
Haizhong Zheng, Yang Zhou, Brian~R Bartoldson, Bhavya Kailkhura, Fan Lai,
  Jiawei Zhao, and Beidi Chen.
\newblock Act only when it pays: Efficient reinforcement learning for llm
  reasoning via selective rollouts.
\newblock \emph{arXiv preprint arXiv:2506.02177}, 2025.

\end{thebibliography}
\bibliographystyle{iclr2026_conference}

\clearpage

\appendix
\section*{Appendix}

\section{Pass Rate and Theoretical Efficiency of Gradient Updates}
\label{appendix:px_theory}

It is shown that the maximum optimization step is bounded by the variance of the reward signal and maximizes at medium difficulty questions, where half of the responses are correct (pass rate, or $p_\theta(x)=0.5$). To ground this statement, we follow earlier work \citep{foster2025learningreasonfrontierlearnability} which states we can approximate the expected policy improvement under a learning rate $\beta$ using a first-order Taylor expansion:
\begin{equation}
    \mathbb{E}_{\pi} [J(\theta_{new}) - J(\theta_{old})] \approx \beta \mathbb{E}_{\pi} [\|\nabla_{\theta} J(\theta)\|^2].
\end{equation}
This shows that optimization speed is proportional to the \textit{Gradient Signal Energy} ($\|\nabla J\|^2$), which measures the steepness of the slope of the loss landscape. For advantage-based algorithms, this decomposes into the Fisher Information Trace of the policy and the scale of the advantage function: 
\begin{equation}
    \mathbb{E}_{\pi} [\|\nabla_{\theta} J\|^2] \propto \underbrace{\mathbb{E} \left[ \left\| \sum_{t} \nabla \log \pi(a_t|s_t) \right\|^2 \right]}_{\text{Fisher Information Trace}} \times \underbrace{\mathbb{E}[(A(s, a))^2]}_{\text{Advantage Magnitude}}.
\end{equation}
The Trace of the Fisher Information Matrix (FIM) measures how sensitive a probability distribution is to changes in weights ($\theta$). A high value means a tiny change in weights causes a large shift in output probabilities. For this purpose, we must assume that the Fisher Information Trace is independent of the Advantage Magnitude, an assumption also made in \citep{foster2025learningreasonfrontierlearnability}.

Now, define question difficulty as the model’s accuracy in answering questions. Formally,\begin{equation}
    p_{\theta}(x)=\mathbb{E}_{y\sim\pi_\theta}[r(x,y)].
 \end{equation}
In binary reasoning tasks (which is our case), the reward $r \in \{0, 1\}$ follows a Bernoulli distribution with pass rate $p_\theta(x)$. When the advantage is defined as $A = r - \mu$ (as with \textit{Dr. GRPO}), the magnitude term becomes the reward variance:
    \begin{equation}
        \mathbb{E}[(r - p_\theta(x))^2] = \operatorname{Var}(r) = p_\theta(x)(1 - p_\theta(x)).
    \end{equation}

Note that we ignore the finite-sample correction ($1-\frac{1}{G})$ for simplicity. Combining these results, we can state that the upper bound on the loss reduction is proportional to the variance of the rewards:
\begin{equation}
    |L(\theta_{new}) - L(\theta_{old})| \propto p_\theta(x)(1 - p_\theta(x)).
\end{equation}

This shows that the gradient signal vanishes for deterministic outputs ($p_\theta(x) \to 0$ or $1$) and peaks at $p_\theta(x)=0.5$. In other words, optimal prompts are of medium difficulty, with a 50\% pass rate.

\section{Algorithm}
The full pseudo code is of our algorithm is as follows;
\begin{algorithm}[h]
\caption{GRPO with Prompt Replay}
\label{alg:grpo_prompt_replay}
\begin{algorithmic}[1]
\REQUIRE Dataset $D_{\text{fresh}}$, Policy $\pi_{\theta}$, Reference $\pi_{\theta_{\text{old}}}$, Group size $G$.
\REQUIRE Hyperparameters: $\epsilon$ (max replay fraction), $C$ (cooldown), $R$ (max reuse), $p_{\min}, p_{\max}$ (difficulty bounds).
\REQUIRE Objective params: $\eta, \epsilon_{\text{low}}, \epsilon_{\text{high}}$ (for Eq.~\ref{eq:olmo_grpo}).
\STATE Initialize Replay Buffer $\mathcal{B} \leftarrow \emptyset$
\STATE Initialize prompt metadata (usage count $u_x$, last used step $t_x$) for buffer entries.

\FOR{training step $t = 1, 2, \dots$}
    \STATE \textbf{// Batch Construction (Prompt Replay)}
    \STATE Identify eligible prompts $\mathcal{E}_t = \{ x \in \mathcal{B} \mid t - t_x > C \}$
    \STATE Rank $\mathcal{E}_t$ by priority $|p_{\theta}(x) - 0.5|$ (ascending)
    \STATE Determine replay size $N_{\text{buf}} = \min(\lfloor \epsilon \cdot N \rfloor, |\mathcal{E}_t|)$ and fresh size $N_{\text{fresh}} = N - N_{\text{buf}}$
    \STATE Sample batch $X_{\text{buf}}$ from top-$N_{\text{buf}}$ of $\mathcal{E}_t$
    \STATE Sample batch $X_{\text{fresh}}$ from $D_{\text{fresh}}$
    \STATE Combined batch $X \leftarrow X_{\text{buf}} \cup X_{\text{fresh}}$

    \STATE \textbf{// Rollout and Advantage Estimation}
    \FOR{each prompt $x \in X$}
        \STATE Generate $G$ responses $\{y_j\}_{j=1}^G \sim \pi_{\theta_{\text{old}}}(\cdot \mid x)$
        \STATE Compute rewards $\{r(x, y_j)\}_{j=1}^G$ S
        \STATE Estimate pass rate $p_{\theta}(x) \leftarrow \frac{1}{G} \sum_{j=1}^G r(x, y_j)$
        \STATE Compute mean reward $\bar{r} \leftarrow \frac{1}{G} \sum_{j=1}^G r(x, y_j)$
        \FOR{$i = 1$ to $G$}
            \STATE Compute advantage $\hat{A}_i \leftarrow r(x, y_i) - \bar{r}$ \quad \COMMENT{Eq.~\ref{eq:advantage_centered}}
        \ENDFOR
        
        \STATE \textbf{// Buffer Update}
        \STATE Update usage: if $x \in X_{\text{buf}}$, $u_x \leftarrow u_x + 1$
        \STATE Update last used step: $t_x \leftarrow t$
        \IF{$p_{\theta}(x) \in [p_{\min}, p_{\max}]$ \AND ($x \notin \mathcal{B}$ \OR $u_x < R$)}
            \STATE Add/Update $x$ in $\mathcal{B}$ with new $p_{\theta}(x)$
        \ELSE
            \STATE Remove $x$ from $\mathcal{B}$ (if exists)
        \ENDIF
    \ENDFOR

    \STATE \textbf{// Policy Optimization}
    \STATE Compute token-level ratios $\rho_{i,t}(\theta)$ and clipped terms using Eq.~\ref{eq:olmo_grpo}
    \STATE Update $\theta$ by maximizing $J(\theta)$:
    $$
    \nabla_\theta J(\theta) \approx \nabla_\theta \left[ \frac{1}{\sum T_i} \sum_{i,t} \min\left(\frac{\pi_\theta}{\pi_{\theta_{\text{old}}}}, \eta\right) \min\left(\rho_{i,t}\hat{A}_i, \text{clip}(\dots)\hat{A}_i\right) \right]
    $$
    \STATE $\pi_{\theta_{\text{old}}} \leftarrow \pi_{\theta}$
\ENDFOR
\end{algorithmic}
\end{algorithm}

\clearpage

\section{Full Hyperparameters Details}
\label{hpo_appendix}

\begin{table}[h!]
\centering
\caption{Experiment Configuration Summary}
\label{tab:experiment_config}
\definecolor{tablerowgray}{RGB}{255,248,240}
\begin{tabular}{@{}lccc@{}}
\toprule
& \textbf{Qwen3-8B Polaris} & \textbf{Llama-3.2-3B Dolci} & \textbf{Qwen3-8B Dolci} \\
\midrule
\rowcolor{tablerowgray}
Batch size & 32  & 32  & 32 \\
Rollouts per Batch & 16 & 16 & 16 \\
\rowcolor{tablerowgray}
Prompt replay fraction & 0.75 & 0.75 & 0.75 \\
Prompt replay max reuse & 15 & 15 & 15 \\
\rowcolor{tablerowgray}
Prompt replay min pass rate & 0.25 & 0.25 & 0.25 \\
Prompt replay max pass rate & 0.75 & 0.75 & 0.75 \\
\rowcolor{tablerowgray}
Replay cooldown steps & 10 & 10 & 10 \\
Seed & 123 & 123 & 123 \\
\rowcolor{tablerowgray}
Pack length & 12,288 & 8,192 & 8,192 \\
Response length & 11,264 & 7,168 & 7,168 \\
\rowcolor{tablerowgray}
Max prompt length & 1,024 & 1,024 & 1,024 \\
vLLM engines & 4 & 4 & 4 \\
\rowcolor{tablerowgray}
Tensor parallel (TP) & 2 & 1 & 1 \\
Num learners & 4 & 4 & 4 \\
\rowcolor{tablerowgray}
Samples per prompt & 16 & 16 & 16 \\
Unique prompts per rollout & 32 & 32 & 32 \\
\rowcolor{tablerowgray}
Learning rate & $1.0 \times 10^{-6}$ & $1.0 \times 10^{-6}$ & $1.0 \times 10^{-6}$ \\
Sampling temperature & 1 & 1 & 1 \\
\bottomrule
\label{HPs}
\end{tabular}
\end{table}

\section{Supplementary Results on the Main Runs: Steps vs Time, Sequence Length, Reward, Entropy}

\begin{figure}[h]
    \centering
    \makebox[\linewidth][c]{%
    \resizebox{1.3\linewidth}{!}{%
        \begin{minipage}{\linewidth}
            \centering

            \begin{subfigure}[b]{0.32\linewidth}
                \centering
                \includegraphics[width=1\linewidth]{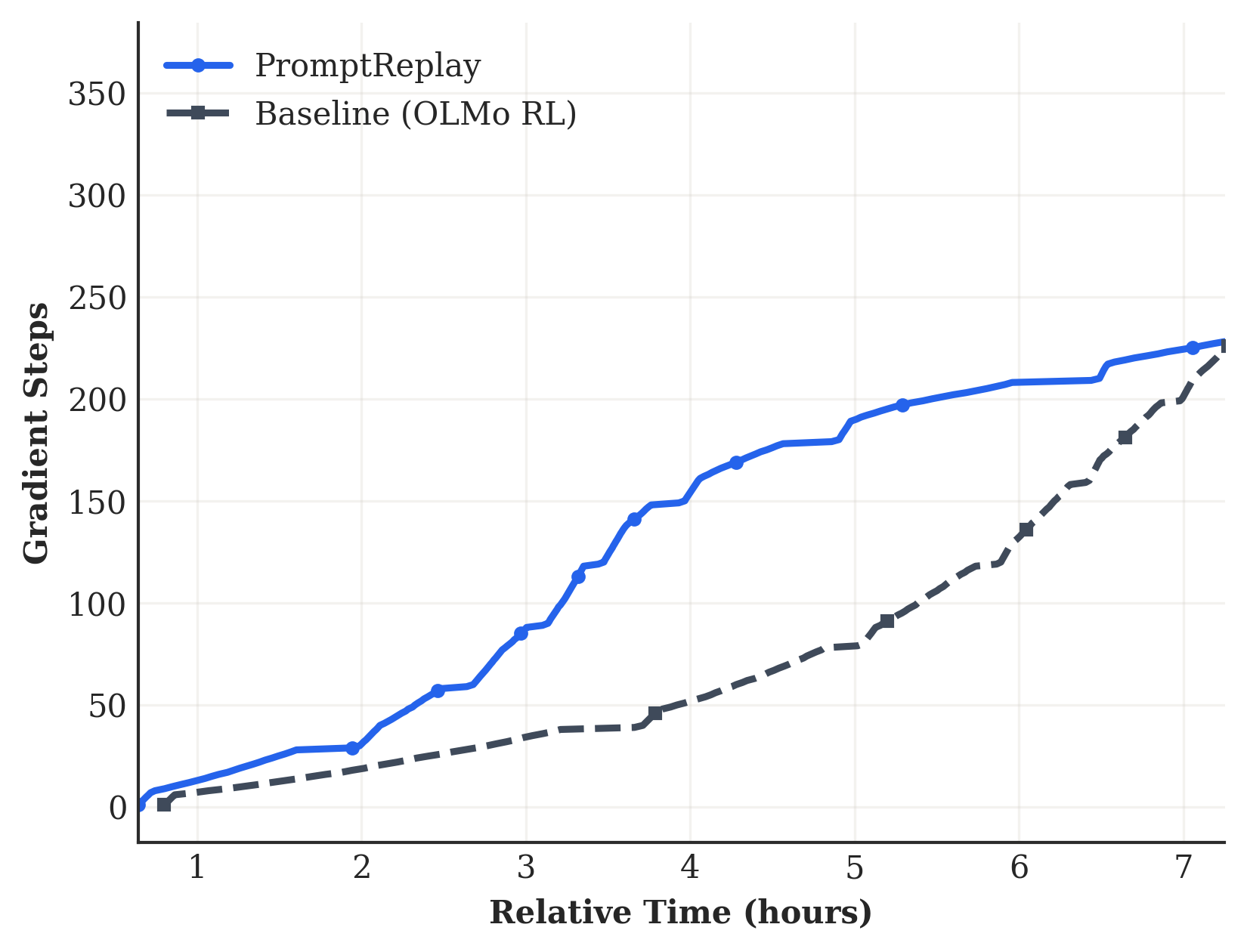}
                \caption{Llama 3.1 3B Dolci}
                \label{fig:steps_llama}
            \end{subfigure}\hfill
            \begin{subfigure}[b]{0.32\linewidth}
                \centering
                \includegraphics[width=1\linewidth]{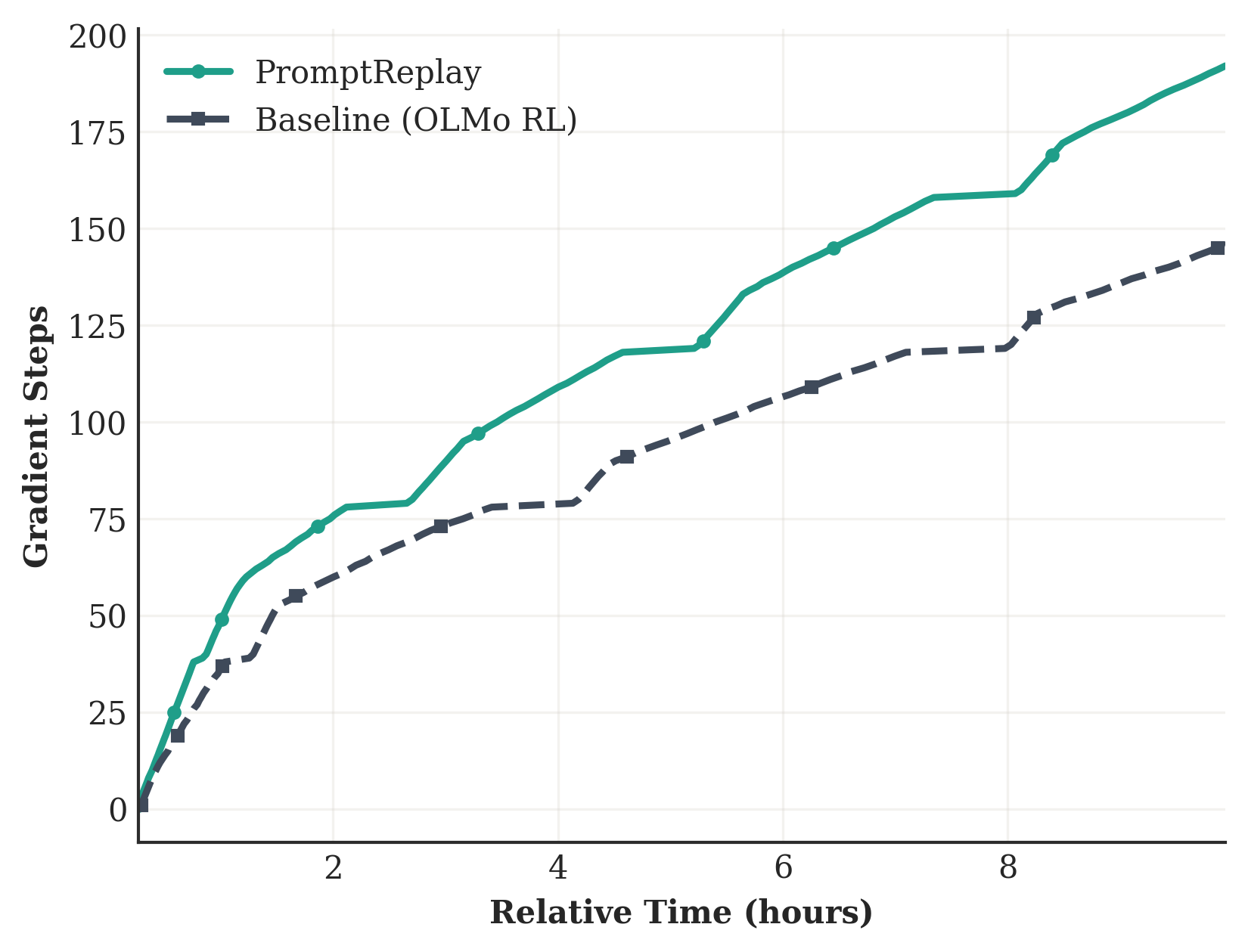}
                \caption{Qwen 3 8B Dolci}
                \label{fig:steps_qwen_dolci}
            \end{subfigure}\hfill
            \begin{subfigure}[b]{0.32\linewidth}
                \centering
                \includegraphics[width=1\linewidth]{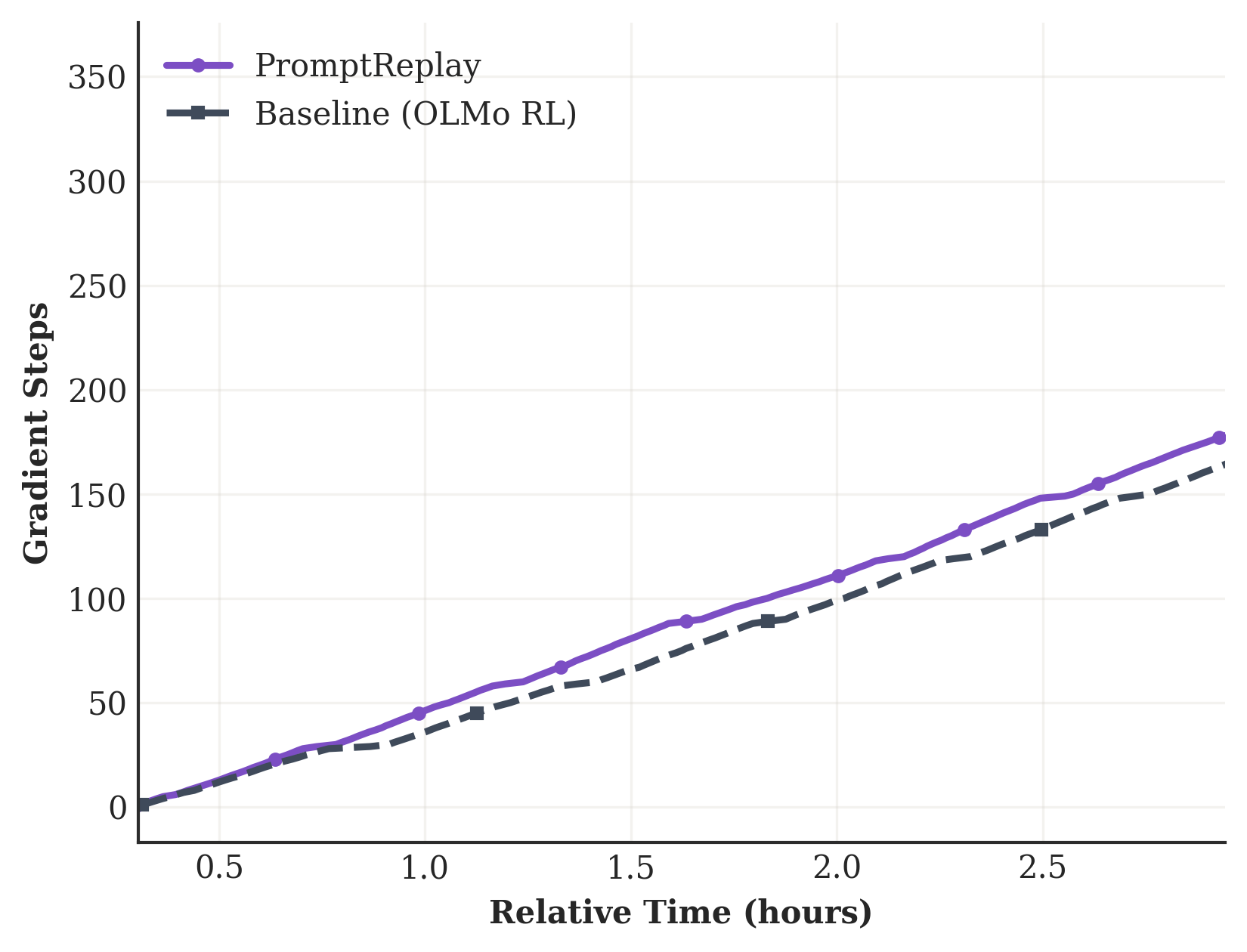}
                \caption{Qwen 3 8B Polaris}
                \label{fig:steps_qwen_polaris}
            \end{subfigure}

            \vspace{0.7em}

            \begin{subfigure}[b]{0.32\linewidth}
                \centering
                \includegraphics[width=1\linewidth]{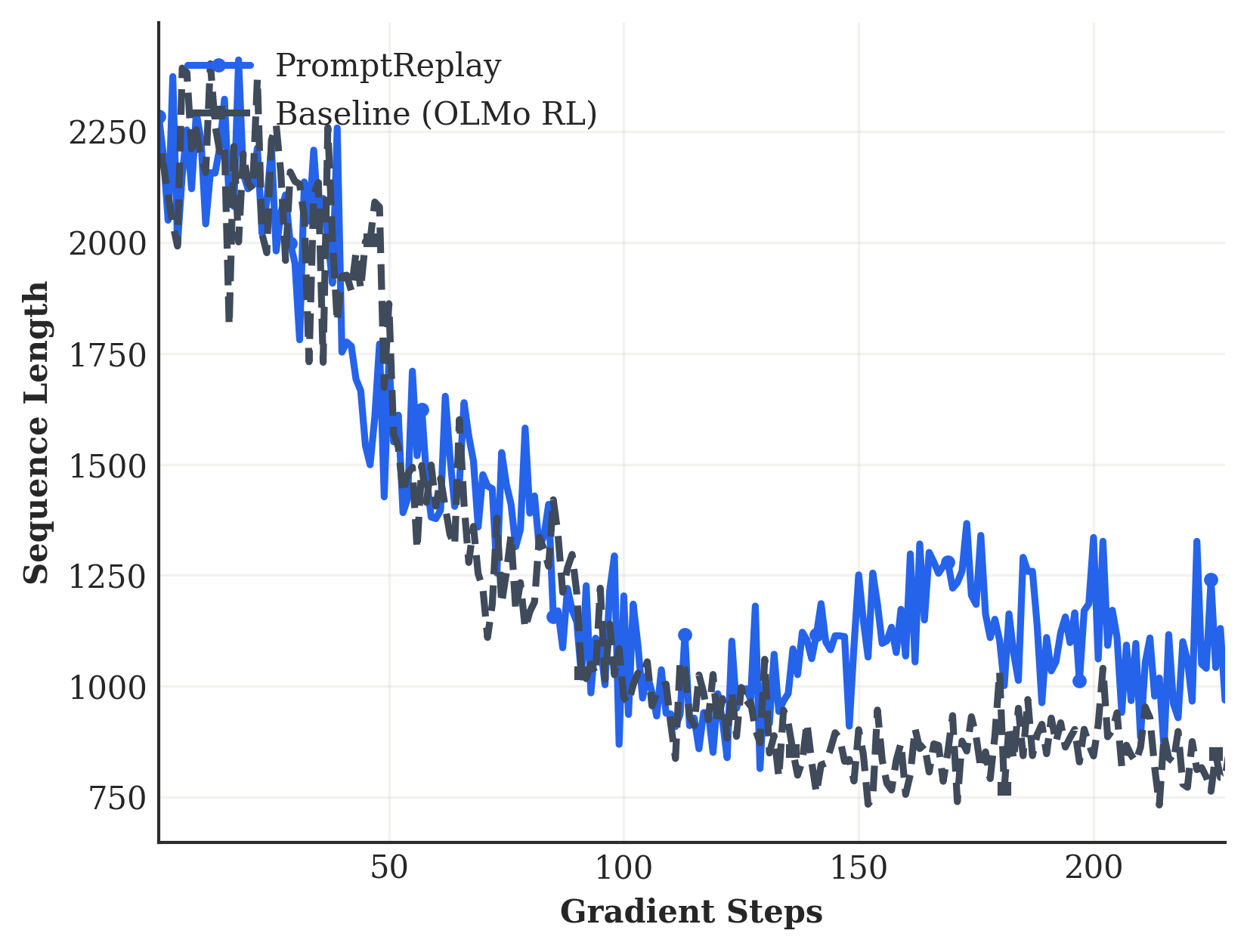}
                \caption{Llama 3.1 3B (Dolci)}
                \label{fig:seqlen_llama}
            \end{subfigure}\hfill
            \begin{subfigure}[b]{0.32\linewidth}
                \centering
                \includegraphics[width=1\linewidth]{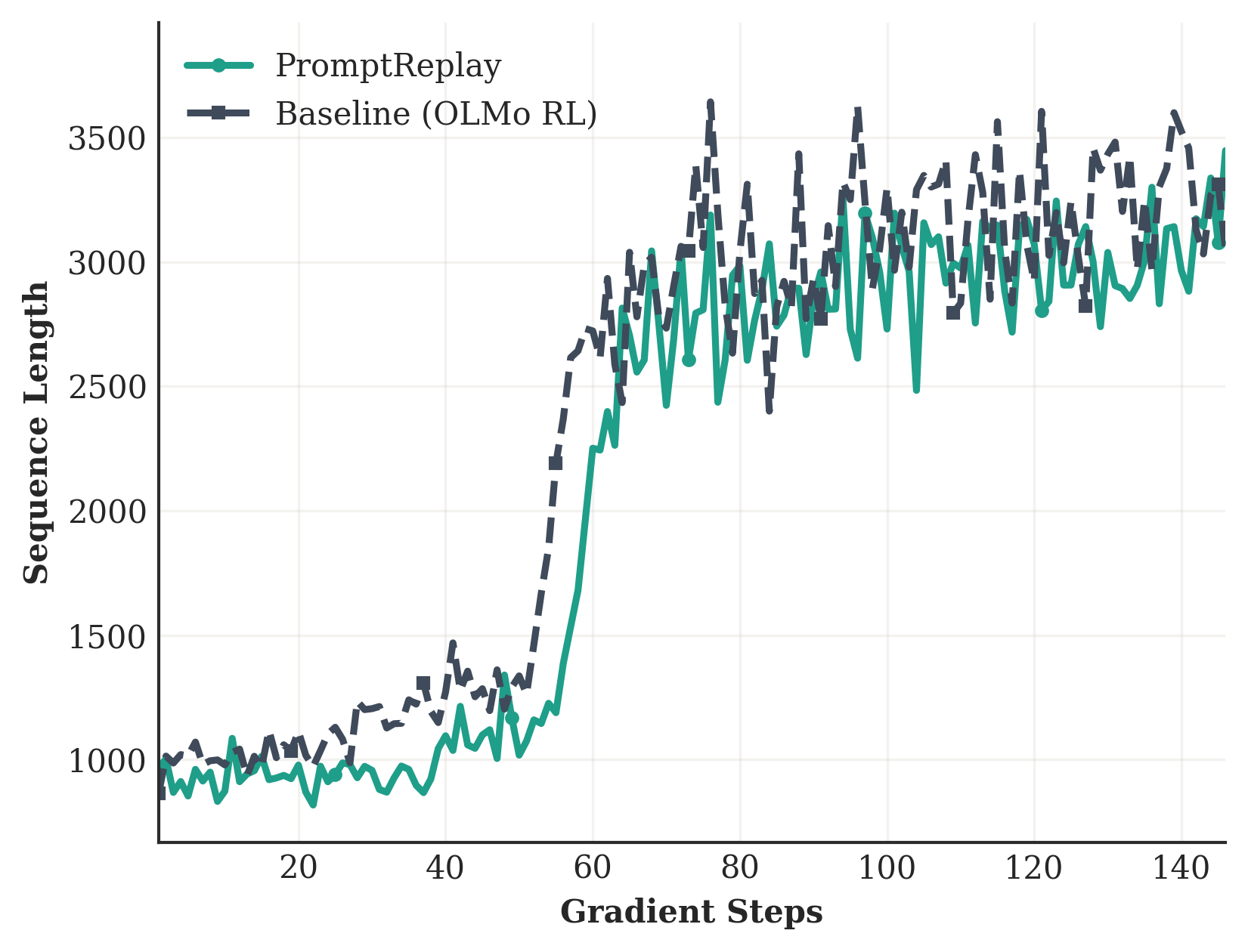}
                \caption{Qwen 3 8B Dolci}
                \label{fig:seqlen_qwen_dolci}
            \end{subfigure}\hfill
            \begin{subfigure}[b]{0.32\linewidth}
                \centering
                \includegraphics[width=1\linewidth]{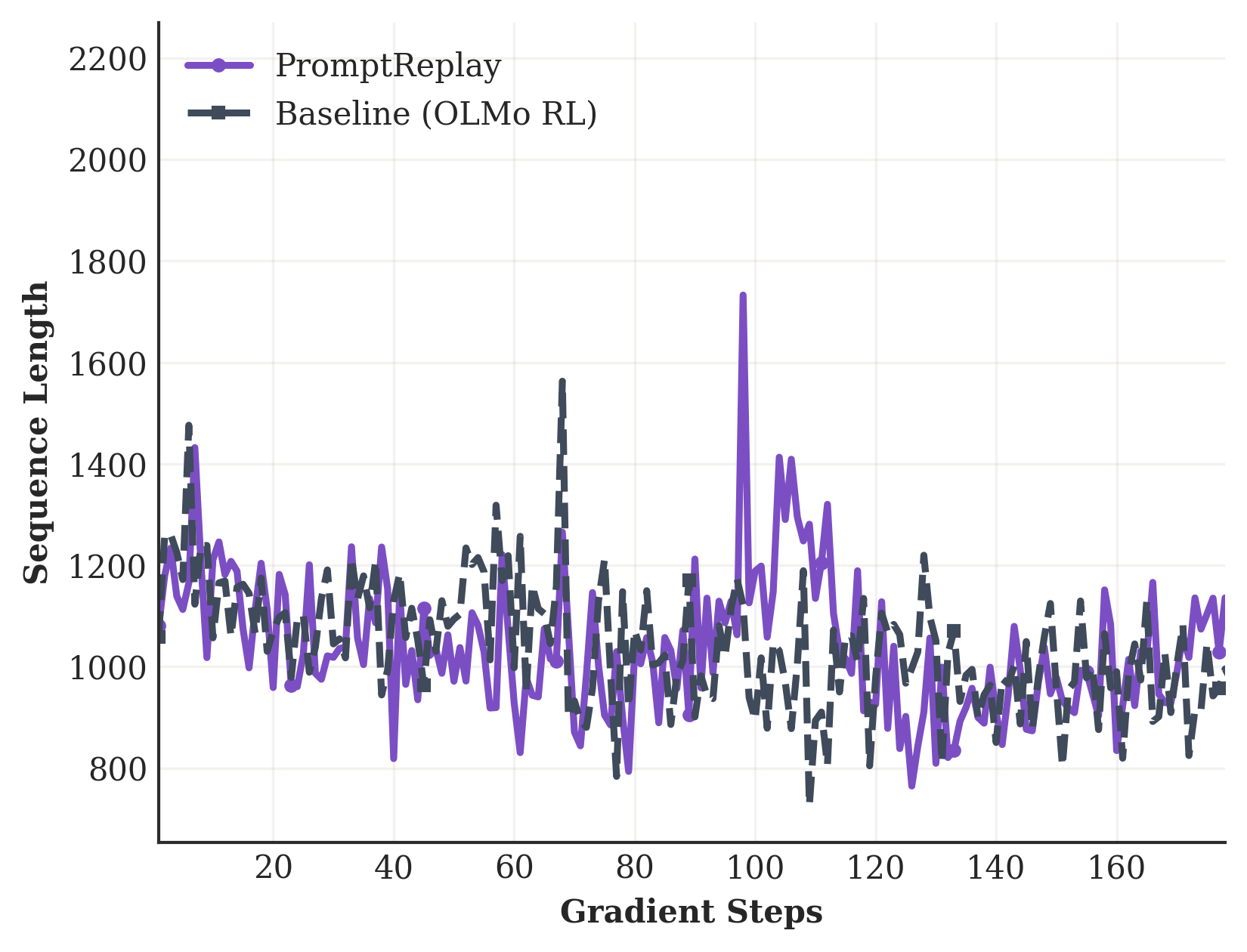}
                \caption{Qwen 3 8B Polaris}
                \label{fig:seqlen_qwen_polaris}
            \end{subfigure}

            \vspace{0.7em}

            \begin{subfigure}[b]{0.32\linewidth}
                \centering
                \includegraphics[width=1\linewidth]{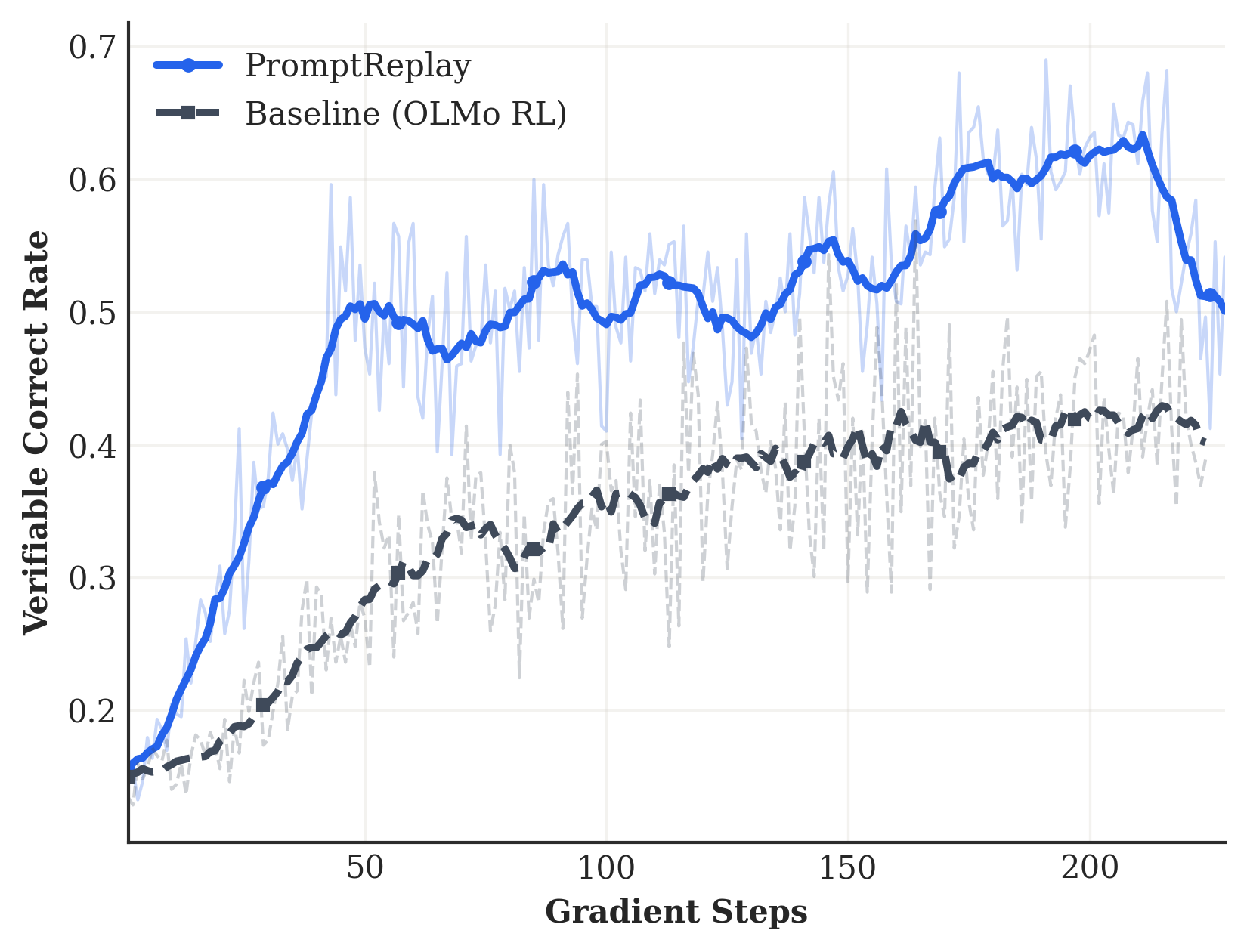}
                \caption{Llama 3.1 3B (Dolci)}
                \label{fig:reward_llama}
            \end{subfigure}\hfill
            \begin{subfigure}[b]{0.32\linewidth}
                \centering
                \includegraphics[width=1\linewidth]{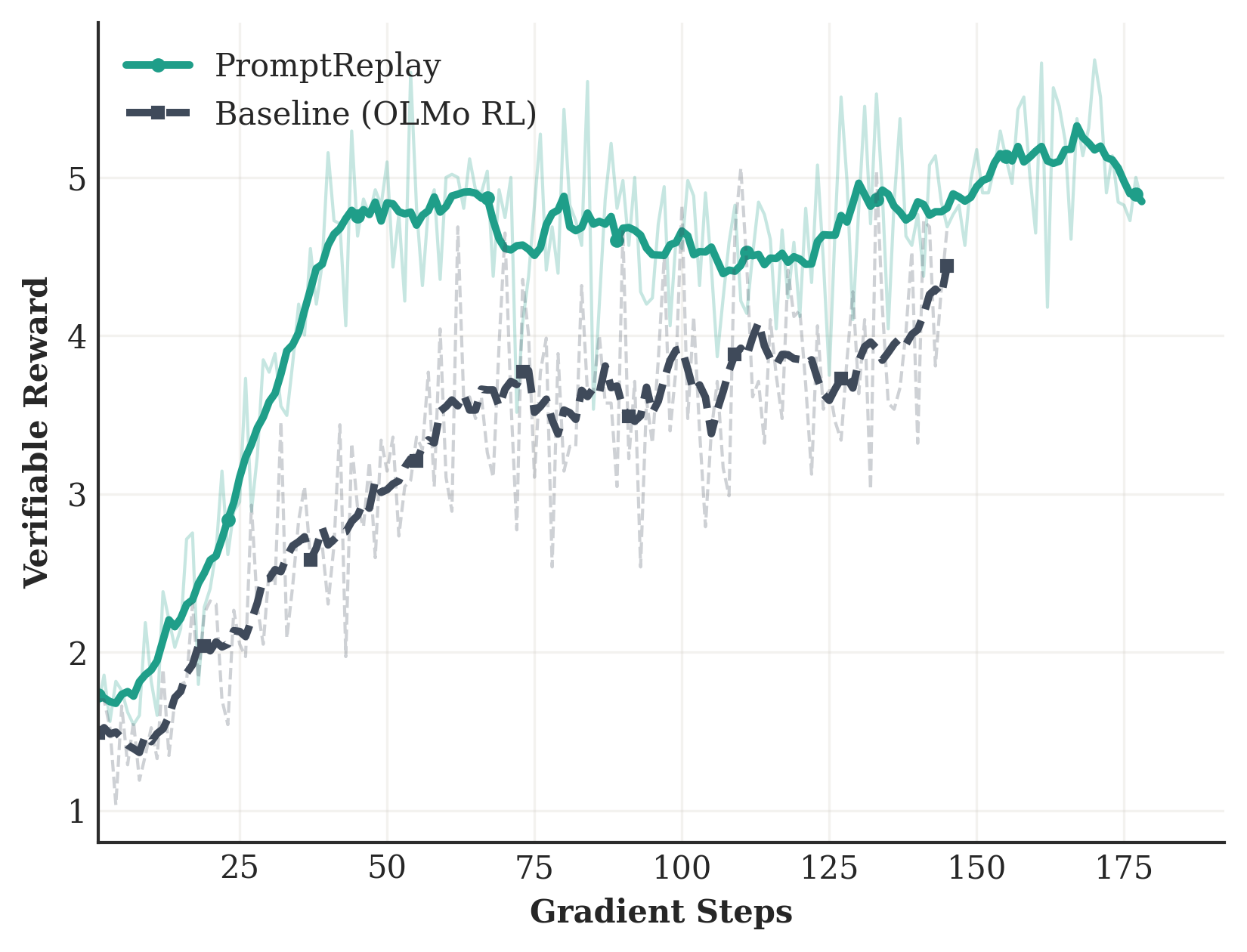}
                \caption{Qwen 3 8B Dolci}
                \label{fig:reward_qwen_dolci}
            \end{subfigure}\hfill
            \begin{subfigure}[b]{0.32\linewidth}
                \centering
                \includegraphics[width=1\linewidth]{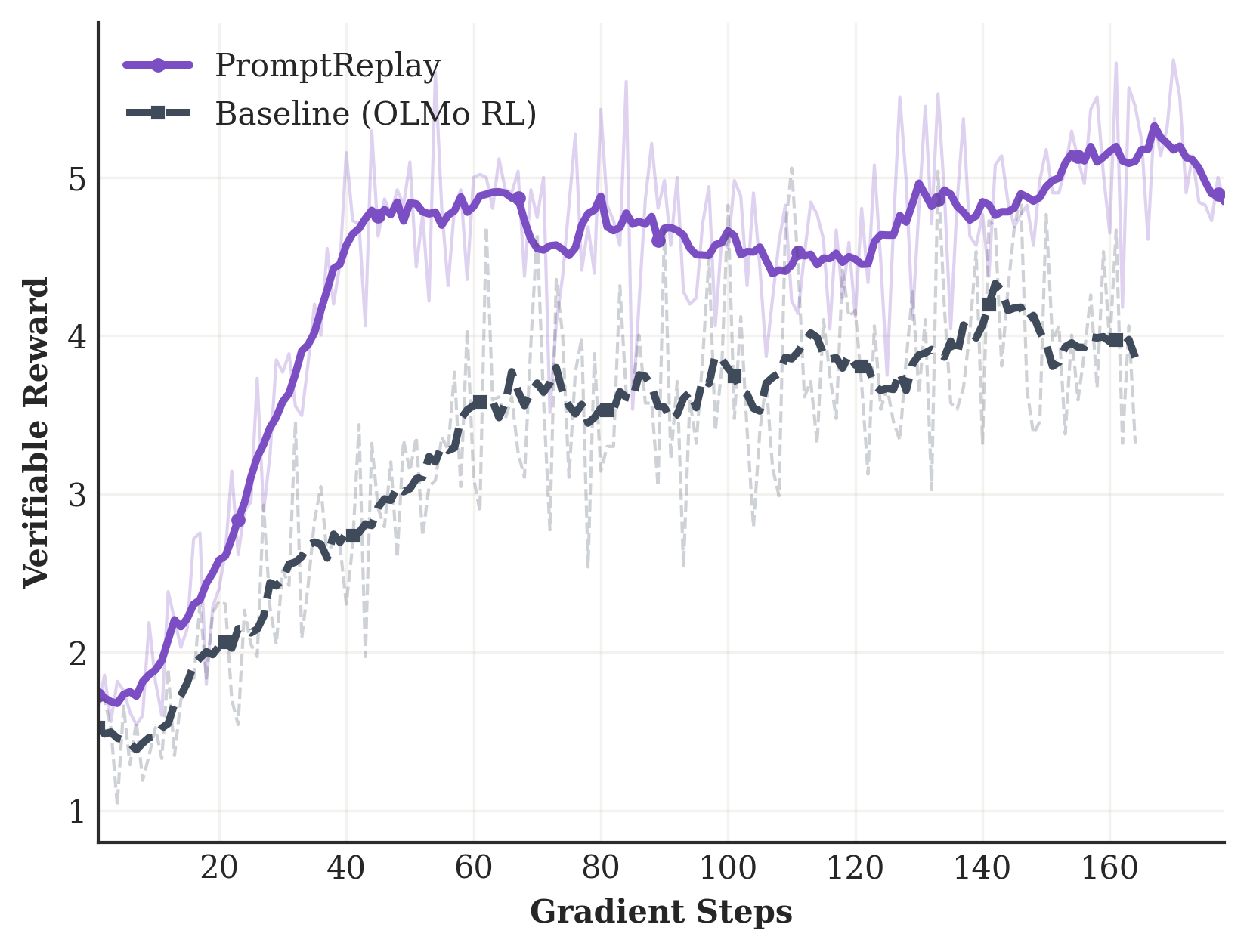}
                \caption{Qwen 3 8B Polaris}
                \label{fig:reward_qwen_polaris}
            \end{subfigure}

            \vspace{0.7em}

            \begin{subfigure}[b]{0.32\linewidth}
                \centering
                \includegraphics[width=1\linewidth]{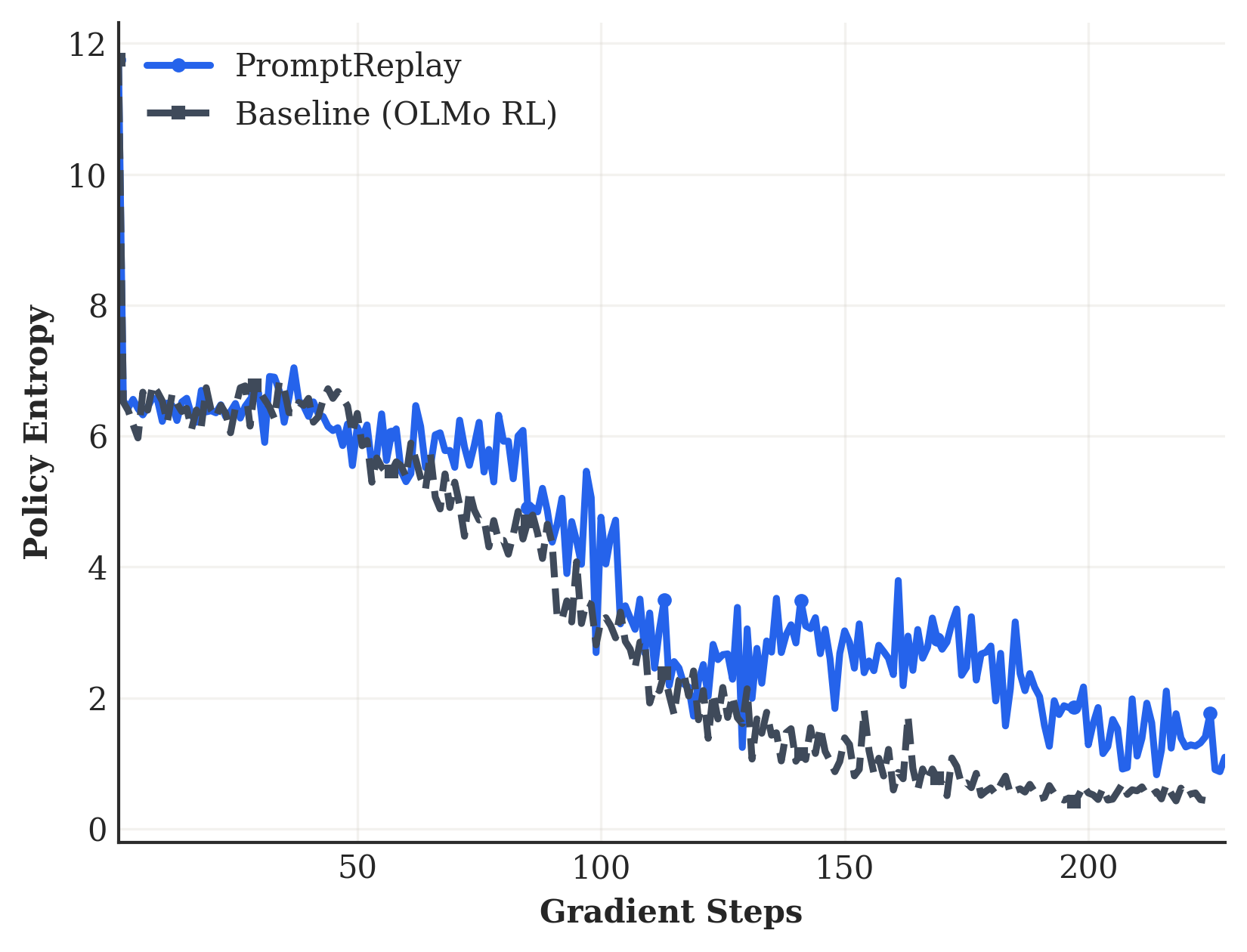}
                \caption{Llama 3.1 3B (Dolci)}
                \label{fig:entropy_llama}
            \end{subfigure}\hfill
            \begin{subfigure}[b]{0.32\linewidth}
                \centering
                \includegraphics[width=1\linewidth]{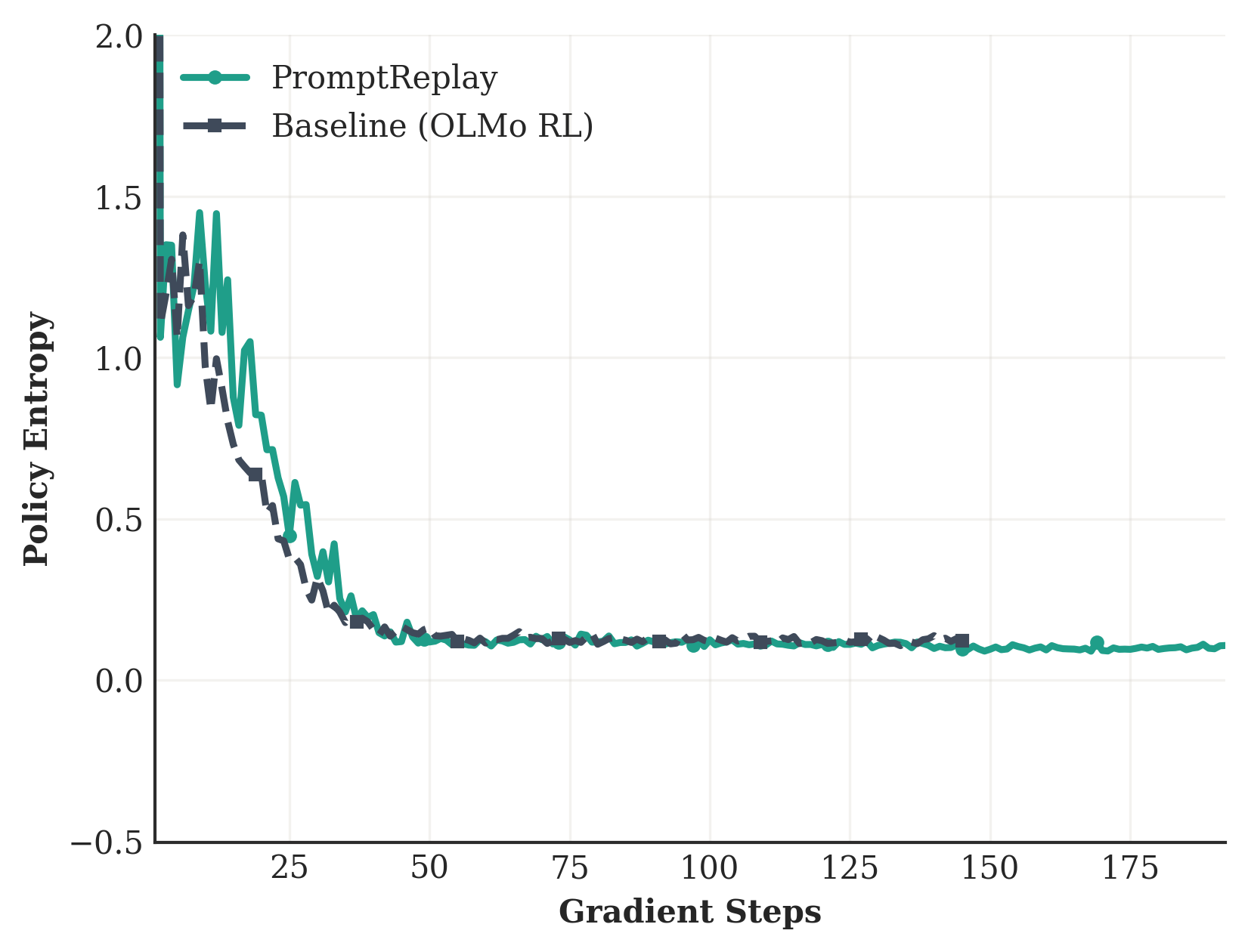}
                \caption{Qwen 3 8B Dolci}
                \label{fig:entropy_qwen_dolci}
            \end{subfigure}\hfill
            \begin{subfigure}[b]{0.32\linewidth}
                \centering
                \includegraphics[width=1\linewidth]{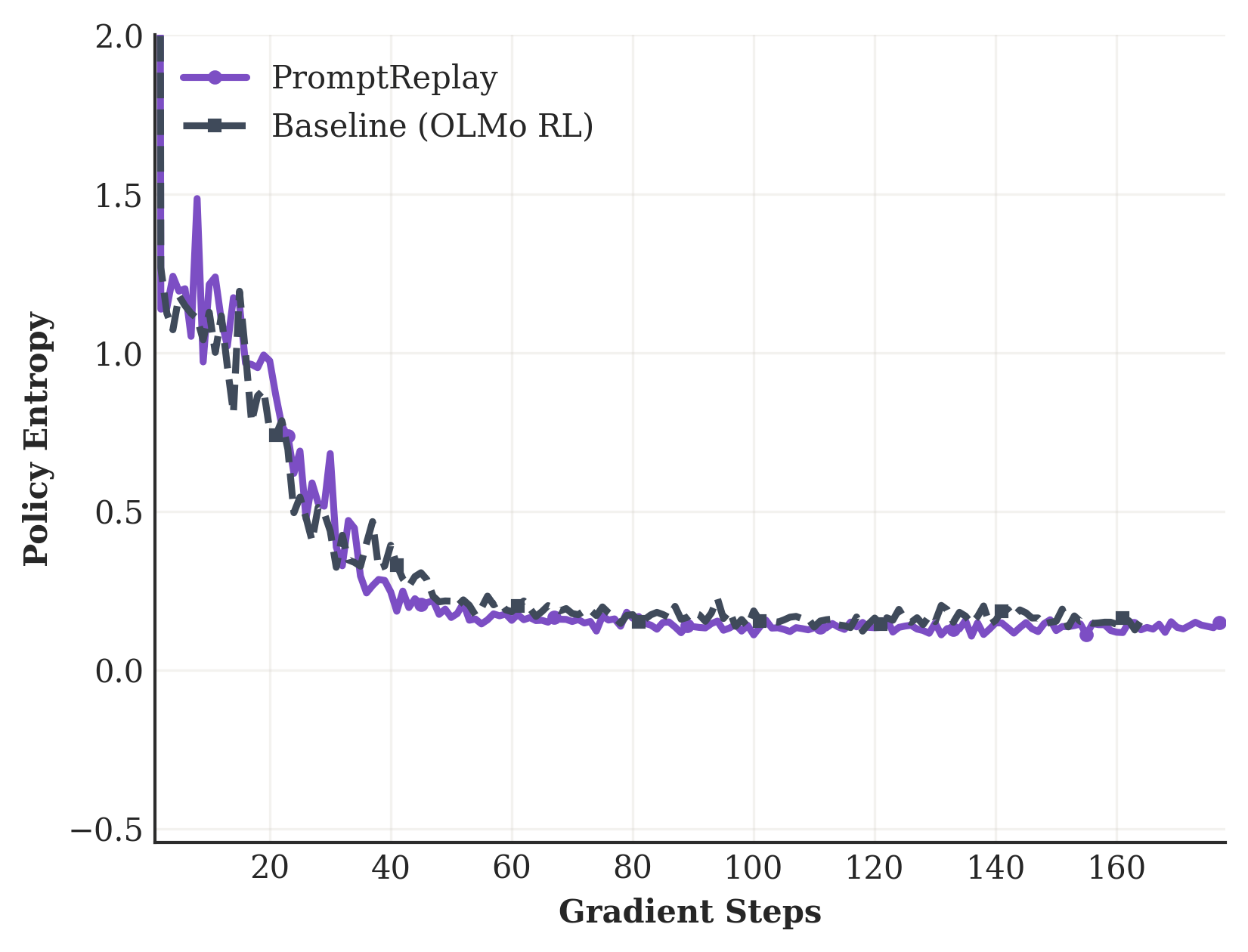}
                \caption{Qwen 3 8B Polaris}
                \label{fig:entropy_qwen_polaris}
            \end{subfigure}

        \end{minipage}%
    }}%
    \caption{Training dynamics for the main results: (row 1) steps vs.\ time, (row 2) sequence length, (row 3) verifiable reward, (row 4) policy entropy. Columns correspond to Llama 3.1 3B (Dolci), Qwen 3 8B (Dolci), and Qwen 3 8B (Polaris).}
    \label{fig:sup_results}
\end{figure}

For \ref{fig:sup_results} a) when Prompt Replay takes effect, Llama shows more steps than the baseline 150 training steps in 4 fours compared to 50 from baseline - this was effect was seen in the avg benchmark accuracy yet it plateaus. Investigating the plateau of steps Prompt Replay, logs show a 20 minute for a single step where the usual time was 1-2 minute. This could be partially explained by the increase in sequence length, but likely bugs appeared at the infrastructure which stalled the run, the real cause remain unknown. 

Qwen 3 8B trained on Dolci dataset (Fig. \ref{fig:sup_results} b))shows the expected behaviour where prompt replay trains faster, and the gap increases over time, where at the end of training, Prompt replay reach about 30\% more steps.

Qwen 3 8B trained on Polaris dataset (Fig. \ref{fig:sup_results} c)) shows the steps over time are very similar, the reason being that the run was given twice the GPU count for rollouts (thus generating responses was no longer the bottleneck), even though Prompt Replay has less rollouts wasted on zero variance prompts. This was unexpected, as its sequence length (Fig \ref{fig:sup_results} f) ) remains fairly constant at around 1000-1200 tokens. It was expected that the sequence length to have a sharp incline to half of the allowed answer length, that is around 6000 tokens. That can be seen is happening for the same model train on Dolci dataset in Fig. \ref{fig:sup_results} e). 

We can see in the Fig. \ref{fig:sup_results} k) and l) that the Qwen model has a sharper entropy decline, reaching much lower values than Llama, reducing exploration.

\end{document}